\documentclass[11pt]{article}

\usepackage[]{acl}

\usepackage{times}
\usepackage{latexsym}

\usepackage[T1]{fontenc}

\usepackage[utf8]{inputenc}

\usepackage{microtype}

\usepackage{inconsolata}

\usepackage{graphicx}

\usepackage{tikz}
\usetikzlibrary{shapes.geometric, arrows, positioning, fit, calc}
\usepackage{amsmath}
\usepackage{filecontents}
\usepackage[table]{xcolor}
\usepackage{booktabs}
\usepackage{graphicx}
\usepackage{amssymb}

\usepackage{colortbl}
\usepackage{pifont}

\newcommand{\cmark}{\ding{51}}  
\newcommand{\xmark}{\ding{55}}  
\newcommand{\warn}{\textcolor{orange}{\large$\blacktriangle$}}
 
\definecolor{minor}{RGB}{204,255,204}
\definecolor{moderate}{RGB}{255,242,204}
\definecolor{high}{RGB}{255,179,186}

\title{NLP Privacy Risk Identification in Social Media (NLP-PRISM): A Survey}

\author{Dhiman Goswami, Jai Kruthunz Naveen Kumar, Sanchari Das \\
        George Mason University \\ Fairfax, VA, USA\\
        \texttt{\{dgoswam,jnaveenk,sdas35\}@gmu.edu}}

\begin{document}
\maketitle
\begin{abstract}
Natural Language Processing (NLP) is integral to social media analytics but often processes content containing Personally Identifiable Information (PII), behavioral cues, and metadata raising privacy risks such as surveillance, profiling, and targeted advertising. To systematically assess these risks, we review $203$ peer-reviewed papers and propose the~\textit{NLP Privacy Risk Identification in Social Media (NLP-PRISM)} framework, which evaluates vulnerabilities across six dimensions: data collection, preprocessing, visibility, fairness, computational risk, and regulatory compliance. Our analysis shows that transformer models achieve F1-scores ranging from $0.58$–$0.84$, but incur a $1\% - 23\%$ drop under privacy-preserving fine-tuning. Using NLP-PRISM, we examine privacy coverage in six NLP tasks: sentiment analysis ($16$), emotion detection ($14$), offensive language identification ($19$), code-mixed processing ($39$), native language identification ($29$), and dialect detection ($24$) revealing substantial gaps in privacy research. We further found a ($\downarrow 2\%-9\%$) trade-off in model utility, MIA AUC (membership inference attacks) $0.81$, AIA accuracy $0.75$ (attribute inference attacks). Finally, we advocate for stronger anonymization, privacy-aware learning, and fairness-driven training to enable ethical NLP in social media contexts.
\end{abstract}

\section{Introduction}
Social media platforms such as X (formerly Twitter), Facebook, and Reddit generate vast volumes of user-generated content~\cite{luca2015user,naveen2026privacy,gupta2024really,noman2019techies,das2021does,das2020change}, providing valuable resources for natural language processing (NLP)~\cite{adhikari2023evolution,adhikaripolicypulse,adhikari2022privacy}. However, this content's informal, multilingual, and noisy nature characterized by slang, emojis, abbreviations, and code-mixing poses challenges for model robustness and generalization~\cite{gharehchopogh2011analysis,singh2017importance,bioglio2022analysis,cho2020privacy}. Core NLP tasks commonly applied to social media include sentiment analysis~\cite{muhammad-etal-2023-semeval,barnes-etal-2022-semeval,patwa2020sentimix,yue2019survey,xu2025privacy}, emotion detection~\cite{giorgi-etal-2024-findings,mohammad-bravo-marquez-2017-wassa,andalibi2020human,ortiz2023implications}, offensive language identification~\cite{zampieri-etal-2019-semeval,zampieri-etal-2020-semeval,ataei2022pars}, code-mixed processing~\cite{calcs2018shtask,sravani-etal-2021-political,yong-etal-2023-prompting,das2013code,roy2025ensuring}, native language identification~\cite{tetreault2013report,malmasi2017report}, and dialect detection~\cite{malmasi-etal-2016-discriminating,zampieri-etal-2019-report,gaman-etal-2020-report,chifu-etal-2024-vardial}. 

These tasks enable applications such as opinion mining~\cite{sharma2020role}, emotion inference~\cite{canales2014emotion,gill2008language}, hate speech moderation~\cite{davidson2017automated,hounsel2018automatically,he2024you,vishwamitra2024moderating,xu2024characterization}, multilingual communication~\cite{dougruoz2021survey}, second language acquisition~\cite{malmasi2016native,stokes2023language,reitmaier2022opportunities,buschek2021impact,strengers2020adhering}, and sociolinguistic analysis~\cite{faisal-etal-2024-dialectbench,eleta2012multilingual}. While large language models have advanced these tasks, they also raise concerns about interpretability and user privacy~\cite{meier2024llm,silva2022privacy,adhikari2025natural}.

Privacy risks in social media NLP arise from adversaries such as platform operators, scrapers, or state actors capable of accessing raw data or models to perform profiling, de-anonymization, or inference attacks~\cite{beigi2020survey,zhang2018privacy}. Given the prevalence of personally identifiable information (PII), geolocation, and behavioral cues in social media text~\cite{lucas2008flybynight}, such data is highly vulnerable to surveillance. Moreover, large models have been shown to memorize sensitive information during training~\cite{das2024security,pan2020privacy}, prompting privacy-preserving efforts such as differential privacy (DP), federated learning (FL), and data anonymization~\cite{bonneau2009privacy,adu2008social,mondal2014understanding,moore2024negative}. DP injects training noise to protect user identity~\cite{wang2017protecting}, while FL decentralizes learning to reduce data leakage~\cite{mistry2024federated,khalil2024federated,li2020review}. However, these techniques remain sparsely adopted in real-world NLP systems. Although prior works (e.g.,~\cite{mahendran2021privacy}) discuss general privacy in NLP, task-specific analyses for social media contexts are limited.

To address this gap, we introduce \textbf{NLP Privacy Risk Identification in Social Media (NLP-PRISM)}, a framework for systematically characterizing privacy risks and mitigation strategies across social media NLP tasks. Using NLP-PRISM, we conducted a systematic review of $3{,}982$ papers and selected $203$ peer-reviewed works from major NLP, privacy, and HCI venues, guided by research question: \emph{What specific privacy risks emerge when key NLP tasks such as sentiment analysis, emotion detection, offensive language identification, code-mixed text processing, native language identification, and dialect detection are applied to social media data, where models may inadvertently reveal or infer sensitive personal attributes, user identities, or demographic information?}

\noindent \textbf{Key Contributions:} 

\noindent (1) We present the first comprehensive systematization of privacy vulnerabilities in social media NLP, analyzing $203$ peer-reviewed studies across six core tasks. Our analysis identifies previously underexplored threats, including latent user re-identification, attribute inference, and representation-level demographic leakage. 

\noindent (2) We propose \textit{NLP-PRISM}, a six-dimensional framework that captures privacy risks across data collection and usage, preprocessing and anonymization, visibility and profiling, computational vulnerabilities, bias, fairness and discrimination, and regulatory and ethical AI considerations. 

\noindent (3) We conduct a quantitative assessment of the privacy utility trade-off using transformers (XLM-R, GPT-2, and FLAN-T5) combined with privacy-preserving interventions such as named-entity masking, text perturbation, and noise addition. Adversarial evaluations using membership inference (MIA) and attribute inference (AIA) attacks demonstrate significant information leakage (MIA AUC up to $0.81$ and AIA accuracy up to $0.75$) and measurable performance degradation (F1 score decreases ranging from $1\%$ to $23\%$), with identity-centric tasks showing the highest vulnerability.

\section{NLP-PRISM: Privacy Risk Framework}
\label{sec:NLP}
\begin{figure*}[!ht]
\centering
\includegraphics[width=\linewidth]{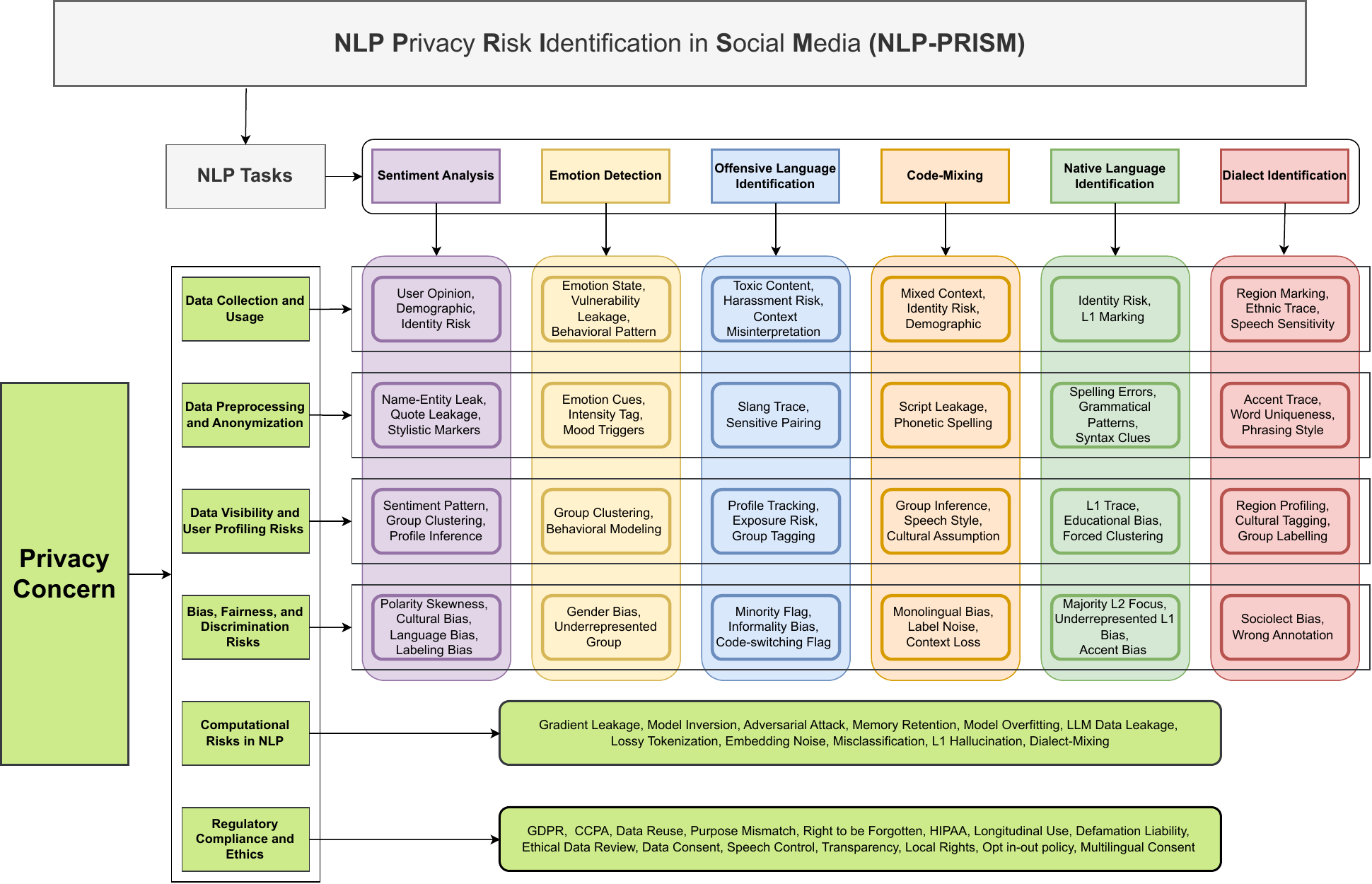}
\caption{NLP-PRISM Framework Highlighting Privacy Risks in NLP-based Social Media Applications}
\label{fig:NLP-PRISM}
\end{figure*}

The~\emph{NLP Privacy Risk Identification in Social Media (NLP-PRISM)} framework offers a comprehensive, multi-dimensional view of privacy-aware NLP by analyzing vulnerabilities, computational risks, and mitigation strategies across the NLP lifecycle. Built upon six interrelated dimensions, it adapts and extends the \textit{Cross-Cultural Privacy Framework}~\cite{ur2013cross}, \textit{OVERRIDE}~\cite{raghavan2012override}, and the \textit{Privacy Risk Assessment Framework (PRAF)}~\cite{saka2024evaluating}. The Cross-Cultural Privacy Framework introduces key factors such as cultural norms, user expectations, and legal contexts shaping privacy behaviors like pseudonymity and disclosure. The OVERRIDE model adds context-aware controls adjusting privacy preferences dynamically, a concept vital for social media environments. Finally, PRAF provides a foundation for assessing compliance, usability, and third-party data sharing. Figure~\ref{fig:NLP-PRISM} summarizes the NLP-PRISM architecture.

\subsection{Data Collection and Usage}
This initial step identifies how privacy risks vary across NLP tasks that depend on user-generated content, often collected without explicit consent. Such data can expose identities, behavioral traits, or demographics~\cite{mondal2014understanding}. For instance, sentiment and emotion analysis infer users' moods and mental states, while offensive language identification misclassifies minority speech as toxic, reinforcing social biases. Code-mixed data reveal bilingual or cultural traits, and native language or dialect identification can expose ethnicity or regional origin~\cite{acharya2025tracing}. Thus, NLP-PRISM underscores that collection processes are inherently tied to identity leakage, emphasizing the need for privacy-aware sourcing practices.

\subsection{Data Preprocessing and Anonymization}
Traditional anonymization through entity masking or token replacement often fails to protect against linguistic leakage. Subtle cues like stylistic markers, slang, or phonetic spellings persist, enabling re-identification, particularly in emotion detection and toxicity tasks~\cite{danescu2011mark}. Dialect and native language models retain accent and syntax traces, while code-mixed data preserve transliteration artifacts and language switches. These residual signals can be exploited by deep learning models for inference attacks~\cite{mei2017inference}. NLP-PRISM thus maps these limitations across tasks, showing how linguistic artifacts remain potent carriers of implicit identity cues.

\subsection{Data Visibility and User Profiling Risks}
As preprocessed data enter models, visibility and profiling risks intensify~\cite{yanushkevich2018understanding}. Sentiment and emotion models may infer psychological or political attributes; offensive language models can marginalize minority dialects by labeling them as toxic. Code-mixed data expose bilingual behavior, and dialect identification reveals demographic traits. When these linguistic features are cross-linked with external datasets, they enable user clustering, behavioral profiling, and discrimination~\cite{zafarani2013connecting}. NLP-PRISM highlights this transition as a pivotal point where model visibility amplifies privacy exposure.

\subsection{Bias, Fairness, and Discrimination Risks}
Bias-driven inequities compound privacy harms by reinforcing marginalization~\cite{mai2016marginalization}. Offensive language models often flag dialectal or culturally marked expressions as toxic, while sentiment and emotion models misinterpret cultural affect~\cite{liu2023cultural}. Underrepresented code-mixed data face monolingual bias and label noise, marginalizing minority voices~\cite{raza2025detecting}. Native language and dialect identification can miscluster speakers by ethnic or regional traits. In NLP-PRISM, these discriminatory tendencies are viewed as privacy violations and expose users' social or cultural affiliations without consent and exacerbating reputational risks.

\subsection{Computational Privacy Risks in NLP}
Beyond task-specific issues, NLP architectures face systemic vulnerabilities. Large models fine-tuned on social media text are susceptible to membership inference~\cite{shokri2017membership} and model inversion~\cite{fredrikson2015model} attacks that reconstruct sensitive samples. Features like lossy tokenization, embedding leakage, and gradient tracing may expose training data even in federated setups. Moreover, hallucinations in generative models can reproduce verbatim text, heightening risk. Hence, NLP-PRISM classifies computational risks as persistent threats requiring privacy-preserving training and inference controls.

\subsection{Regulatory Compliance and Ethics}
This dimension examines how legal and ethical standards constrain NLP use. Systems must comply with GDPR~\cite{voigt2017eu}, CCPA~\cite{goldman2020introduction}, and HIPAA, yet enforcement is hindered by opaque data reuse and the inference of unregulated attributes. Sentiment and language identification models may reveal psychological or demographic traits beyond current protections. Ethical lapses arise when model outputs misclassify or stigmatize users, especially in culturally sensitive contexts. Without transparent, opt-in frameworks, NLP deployments risk violating both legal and moral boundaries.

\section{Privacy Risk with Large Language Models (LLMs)}

While the NLP-PRISM framework characterizes privacy risks across core NLP tasks, the adoption of LLMs introduces additional model-centric privacy threats that cut across the six tasks. Unlike task-specific models, LLMs operate as general-purpose systems trained on massive, heterogeneous corpora~\cite{goswami2025multilingual}, often collected with limited transparency or control. Due to their scale, contextual learning, and training dynamics, LLMs are vulnerable to risks such as training data extraction, attribute inference, activation inversion, prompt-based leakage, and user-level inference~\cite{das2025security}, which align with and also extend the dimensions captured by NLP-PRISM.

In LLM-based sentiment analysis, privacy risks extend beyond output-level predictions to the exposure of sensitive training content and subjective user expressions. Since models often process highly personal opinions related to health, politics, or lived experiences, leakage at the model level can directly compromise individual privacy.~\citet{dai-etal-2025-stealing} show that adversaries can recover parts of training data through activation inversion attacks, revealing sentiment-bearing personal text even when raw data is not accessible. In addition, privacy auditing studies by~\citet{meng2025rr} demonstrate that LLMs may reproduce sentiment-specific expressions from training data under targeted prompting, raising concerns about inadvertent disclosure of sensitive opinions or experiences.

Emotion detection by LLMs is particularly vulnerable to attribute inference attacks, where sensitive emotional or psychological attributes can be inferred from model outputs or latent representations. Because emotional states are closely tied to mental health and personal well-being, even indirect inference poses significant privacy harm. Research on recollection and ranking behavior in LLMs by \citet{meng2025rr} shows that emotionally salient content is more likely to be memorized and resurfaced during inference, increasing exposure risks under probing prompts. Furthermore, studies on named-entity and attribute inference by \citet{sutton2025named} demonstrate that even anonymized emotional text can leak private information when processed by LLMs during training or output generation.

For offensive language identification, LLMs amplify privacy risks through their ability to memorize and regenerate toxic, abusive, or sensitive examples encountered during training. Such data often originates from real user interactions and may contain identifiable information or contextual cues. Privacy evaluations of LLMs by \cite{ye2025llms} reveal that memorized offensive content indicates a risk of unintended reproduction of harmful or identifying text during downstream use. Additionally, inference-based attacks show that adversaries may deduce whether specific offensive instances or user-generated content were used during training, posing privacy concerns for moderation systems and content reporting pipelines~\cite{mattern-etal-2023-membership}.

LLMs trained on multilingual and code-mixed data introduce unique privacy risks due to cross-lingual representation leakage. Code-mixed content often reflects individual identity, community membership, or migration history, making leakage particularly sensitive. Activation inversion attacks demonstrate that sensitive mixed-language training samples can be reconstructed from intermediate representations in large models, even in decentralized or collaborative training settings~\cite{song-etal-2025-multilingual}. Moreover, privacy auditing work indicates that multilingual LLMs may inadvertently encode user-specific language-mixing patterns, increasing the risk of re-identification or demographic inference~\cite{kim-etal-2024-pruning}.

Native language identification using LLM embeddings raises concerns related to user-level inference, where an attacker can infer whether a particular user, or data from that user, contributed to model training. Since native language often correlates with nationality, ethnicity, or migration status, such inference can have broader societal implications. Recent studies by \cite{choi-etal-2025-safeguarding} show that LLMs encode subtle linguistic signals that persist even after mitigation attempts, enabling adversaries to infer sensitive language-related attributes beyond the intended task.

Dialect identification tasks further compound privacy risks, as dialectal features often correlate with regional, cultural, or socio-demographic attributes at both individual and community levels. When dialect detection is embedded within LLM pipelines, latent representations may capture fine-grained linguistic markers that users did not explicitly consent to share. Research on LLM privacy leakage by \cite{ye2025llms} shows that latent representations can expose such linguistic patterns, enabling attribute inference or reconstruction attacks even when outputs appear benign.

\section{Methodology}
\begin{figure}[!ht]
\centering
\includegraphics[width=\linewidth]{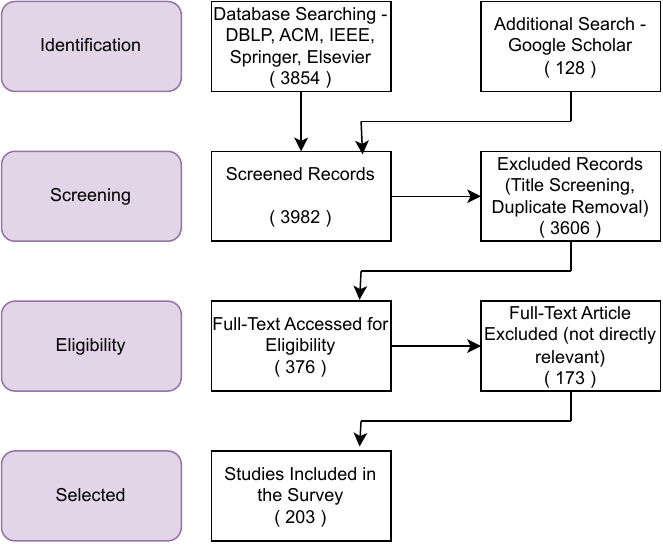}
\caption{Overview of Data Collection : PRISMA.}
\label{fig:prisma}
\vspace{-3mm}
\end{figure}

For this survey, we followed the Preferred Reporting Items for Systematic Reviews and Meta-Analyses (PRISMA) guidelines~\cite{mcinnes2018preferred, moher2010preferred} to ensure a transparent, systematic, and evidence-based process for identifying, screening, and selecting relevant studies. We also adopted the methodological design established in prior works by the last author of this paper~\cite{huang2025systemization,shrestha2022sok,duzgun2022sok,tazi2022sok,das2022sok,zezulak2023sok,tazi2023sok,tazi2022sok,grover2025sok,tazi2024sok,saka2025sok,podapati2025sok,majumdar2021sok,agarwal2025systematic,kishnani2023blockchain,jones2021literature,noah2021exploring,das2019all,shrestha2022exploring}. In addition, our analysis was guided by the NLP-PRISM framework (detailed in Section~\ref{sec:NLP}). Figure~\ref{fig:prisma} provides an overview of the PRISMA-based methodology employed in this study.

\subsection{Paper Identification}
We queried major academic sources that publish high-impact work on NLP, privacy \& security, and HCI, namely \textit{DBLP, ACM Digital Library, IEEE Xplore, Springer,} and \textit{Elsevier}. In addition, we included top-ranked conferences and journals (A*/A venues per CORE ranking\footnote{\url{https://portal.core.edu.au/conf-ranks/}}) and used Google Scholar to capture influential papers not indexed in those databases. The distribution of retrieved items by venue is shown in Appendix Table~\ref{tab:list1}.

\subsection{Paper Screening}
We constructed structured search queries combining general privacy phrases (e.g.,~\textit{social media privacy, privacy concerns in social media, social media privacy risk}) with task-specific terms (e.g.,~\textit{privacy risks in sentiment analysis, emotion detection, offensive language identification, code-mixed text processing, native language identification, dialect identification}). Consequently, our search targeted six core NLP tasks that are especially sensitive in social media contexts: sentiment analysis, emotion detection, offensive language identification, code-mixing, native language identification, and dialect identification. Briefly, these tasks can enable large-scale emotional profiling, deeper mood inferences (including mental-health signals), surveillance or biased moderation, cross-lingual identity leakage, geolocation or demographic inference, and socioeconomic or cultural profiling respectively.

We also included privacy-centric keywords such as \textit{privacy-aware NLP, differential privacy in NLP, federated learning for NLP}, and \textit{anonymization in text processing}. To remain current, the search covered publications up to \textit{December 2024}. Appendix Table~\ref{tab:list2} reports the task-wise paper counts, including privacy-related studies.

\subsection{Paper Exclusion and Final Selection}
Our initial retrieval returned \textbf{3,982} records. We applied a multi-stage filtering pipeline: title/metadata screening to remove duplicates and clearly irrelevant items, abstract-level screening to assess topical relevance to NLP and privacy, and a full-text eligibility review to judge methodological rigor and contributions to privacy-enhancing techniques (e.g., DP, FL, anonymization, adversarial robustness). Two researchers independently conducted the title/abstract and full-text screenings; disagreements were escalated to a third researcher when necessary. After abstract screening, \textbf{3,606} papers were excluded, leaving \textbf{376} for full-text review; a further \textbf{173} were removed on full-text inspection, producing a final set of \textbf{203} studies.

Using the NLP-PRISM framework, we coded each paper across six predefined dimensions: \textit{Data Collection and Usage}, \textit{Data Preprocessing and Anonymization}, \textit{Data Visibility and User Profiling Risks}, \textit{Computational Privacy Risks in NLP}, \textit{Bias, Fairness, and Discrimination Risks}, and \textit{Regulatory Compliance and Ethics}. Two annotators independently labeled a random 25\% subset to assess reliability, yielding an inter-annotator agreement of $87.8\%$ and Cohen's Kappa of $0.756$, which indicates strong reliability as per ~\citet{mchugh2012interrater}.

Finally, to complement our qualitative synthesis, we ran empirical evaluations that quantify privacy leakage under adversarial settings aligned with NLP-PRISM. Specifically, we measured membership inference and attribute inference attacks on fine-tuned transformer models across the six tasks, and we tested privacy interventions such as named-entity masking and noise addition. The task-wise effects on model utility and attack success are summarized in Table~\ref{tab:results}. 

\begin{table*}[!htp]
\centering
\resizebox{\linewidth}{!}{%
\begin{tabular}{p{3.4cm}p{17.5cm}}
\hline
\textbf{Tasks} & \textbf{Privacy Risk}\\
\hline
\textbf{Sentiment Analysis} & 
Data Collection and Usage~\cite{xiao2018mobile,sharma2020role,lohar2021irish},

Data Preprocessing and Anonymization~\cite{lohar2021irish,xiao2018mobile},

Data Visibility and User Profiling Risks~\cite{lohar2021irish,sharma2020role},

Bias, Fairness, and Discrimination Risks~\cite{konate2018sentiment,barbieri2016overview},

Regulatory Compliance and Ethics~\cite{xiao2018mobile,raihan2023sentmix}\\
\hline
\textbf{Emotion Detection} & 
Data Collection and Usage~\cite{raihan2024emomix,teodorescu2023language,arango2024multifold,zampieri-etal-2024-federated},  

Data Preprocessing and Anonymization~\cite{teodorescu2023language},  

Computational Privacy Risks in NLP~\cite{arango2024multifold}, 

Regulatory Compliance and Ethics~\cite{mohamed2022artelingo,raihan2024emomix,teodorescu2023language}\\
\hline
\textbf{Offensive Language Identification} & 
Data Collection and Usage~\cite{zampieri-etal-2024-federated,sigurbergsson-derczynski-2020-offensive,rosenthal2021solid,arango2024multifold,deng-etal-2022-cold}, 

Data Preprocessing and Anonymization~\cite{leonardelli2021agreeing,jeong-etal-2022-kold,sigurbergsson-derczynski-2020-offensive,rosenthal2021solid}, 

Data Visibility and User Profiling Risks~\cite{leonardelli2021agreeing,rosenthal2021solid}, 

Computational Privacy Risks in NLP~\cite{arango2024multifold,xiao-etal-2024-toxicloakcn,morabito2024stop}, 

Regulatory Compliance and Ethics~\cite{zampieri-etal-2024-federated,jeong-etal-2022-kold}\\
\hline
\textbf{Code-Mixing} & 
Data Collection and Usage~\cite{hidayatullah2022systematic},  

Data Preprocessing and Anonymization~\cite{hidayatullah2022systematic},  

Data Visibility and User Profiling Risks~\cite{hidayatullah2022systematic,zhang2023multilingual}, 

Regulatory Compliance and Ethics~\cite{hidayatullah2022systematic,garg2018code}\\
\hline
\textbf{Native Language Identification} & 
Data Collection and Usage~\cite{staicu2023bilingual,bierner2001alternative,goldin2018native,jauhiainen2019automatic},  

Data Preprocessing and Anonymization~\cite{berzak-etal-2017-predicting}, 

Data Visibility and User Profiling Risks~\cite{jauhiainen2019automatic,goldin2018native,aoyama2024modeling}, 

Regulatory Compliance and Ethics~\cite{jauhiainen2019automatic,nguyen2021automatic}\\
\hline
\textbf{Dialect Identification} &  

Data Collection and Usage~\cite{jorgensen2015challenges,huang2015improved,fleisig-etal-2024-linguistic,barot2024tonecheck}, 

Data Preprocessing and Anonymization~\cite{malmasi2015arabic,ferragne2007automatic,mahmud2023cyberbullying}, 

Computational Privacy Risks in NLP~\cite{mahmud2023cyberbullying}, 

Bias, Fairness, and Discrimination Risks~\cite{fleisig-etal-2024-linguistic}, 

Regulatory Compliance and Ethics~\cite{huang2015improved,fleisig-etal-2024-linguistic,faisal-etal-2024-dialectbench,barot2024tonecheck}\\
\hline
\end{tabular}
}
\caption{Survey of Privacy Risks in Social Media NLP Tasks}
\label{tab:list3}
\end{table*}

\section{Results}
Here we provide an overview of the datasets, benchmark competitions, and computational methodologies. Key trends across these components are summarized in Appendix Table~\ref{tab:list4}, with emphasis on their relevance to privacy challenges.

\subsection{Privacy Risks (NLP-PRISM Analysis)}
Table~\ref{tab:list3} summarizes the analyzed papers and the identified privacy risks. A detailed analysis of these risks is provided in Appendix~\ref{appendix:risk}.

\textbf{Data Collection and Usage:} Privacy compliance in data collection and usage remains a critical challenge, yet $19$ of the $203$ papers in our study highlight this. For instance,~\citet{xiao2018mobile} and~\citet{sharma2020role} could improve transparency by clarifying data collection practices in sentiment analysis. ~\citet{lohar2021irish} note privacy issues in public health applications. Likewise,~\citet{raihan2024emomix} and~\citet{teodorescu2023language} could strengthen data policies to prevent inadvertent disclosures. Efforts in offensive language identification, such as~\citet{zampieri-etal-2024-federated} and~\citet{arango2024multifold}, would benefit from stronger control in dataset collection. Similarly,~\citet{sigurbergsson-derczynski-2020-offensive},~\citet{rosenthal2021solid}, and~\citet{deng-etal-2022-cold} could reinforce privacy protections in large-scale curation. Furthermore, while~\citet{hidayatullah2022systematic} and~\citet{bierner2001alternative} contribute valuable code-mixed resources, clearer data usage measures would improve ethical standards. In NLI,~\citet{goldin2018native} and~\citet{huang2015improved} could refine metadata handling to protect user identities. Finally,~\citet{fleisig-etal-2024-linguistic} emphasize compliance when using proprietary models for dialect analysis. Collectively, these studies underline the necessity of robust frameworks for privacy-conscious data collection and usage.

\textbf{Data Preprocessing and Anonymization:} Effective anonymization is vital for privacy protection, yet only $12$ papers offer explicit methods with room for improvement. For example,~\citet{lohar2021irish} and~\citet{xiao2018mobile} could strengthen anonymization by detailing specific steps, while~\citet{teodorescu2023language} and~\citet{leonardelli2021agreeing} could integrate clearer anonymization protocols. Research on offensive language, such as~\citet{jeong-etal-2022-kold} and~\citet{sigurbergsson-derczynski-2020-offensive}, would benefit from improved anonymization during annotation. Similarly,~\citet{hidayatullah2022systematic} and~\citet{malmasi2015arabic} could enhance documentation of anonymization practices, ensuring compliance. Moreover,~\citet{ferragne2007automatic} and~\citet{mahmud2023cyberbullying} provide insights into dialect and offensive language detection, where integrating anonymization would further enhance privacy. Overall, these works highlight the growing awareness of user data protection and the need for transparent anonymization standards.

\textbf{Data Visibility and User Profiling Risks:} Managing data visibility and minimizing profiling risks are crucial for privacy, yet only $4\%$ of the reviewed papers address these issues. For instance,~\citet{sharma2020role} raise concerns about AI-driven sentiment monitoring, underscoring the need for informed consent. Similarly,~\citet{lohar2021irish} and~\citet{leonardelli2021agreeing} discuss risks in sentiment tracking and offensive language classification, calling for clearer data protection mechanisms. In addition,~\citet{lukas2023analyzing} highlight data leakage risks in large-scale models, which demand stricter access control.~\citet{hidayatullah2022systematic} and~\citet{goldin2018native} illustrate how dataset exposure can lead to unintended profiling, advocating privacy-aware dataset structuring. Finally,~\citet{zhang2023multilingual},~\citet{jauhiainen2019automatic}, and~\citet{aoyama2024modeling} emphasize privacy concerns in code-mixed processing and NLI, stressing the need for stronger institutional privacy policies.

\textbf{Computational Privacy Risks in NLP:} Computational privacy concerns arise when data handling, model architectures, or annotation procedures do not fully align with privacy regulations. \textit{We have found this aspect to be addressed in only about $2\%$ papers of our study}.~\citet{arango2024multifold} emphasize the importance of ensuring privacy compliance in cross-domain adaptation. Similarly,~\citet{xiao-etal-2024-toxicloakcn} and~\citet{morabito2024stop} explore the detection of offensive language in large models, and further evaluation of computational privacy risks could refine their approaches.~\citet{mahmud2023cyberbullying} and~\citet{rizwan2020hate} contribute to low-resource language datasets, where clearer annotation and masking strategies would mitigate privacy concerns. 

\begin{table*}[htp]
\centering

\resizebox{\textwidth}{!}{%
\begin{tabular}{l|cccc|cccc|cccc}
\hline
 & \multicolumn{4}{c|}{\textbf{XLM-R}} & \multicolumn{4}{c|}{\textbf{GPT-2}} & \multicolumn{4}{c}{\textbf{FLAN-T5}}\\
\cline{2-13}
\textbf{Tasks} & \textbf{F1} & \textbf{F1(P)} & \textbf{MIA AUC} & \textbf{AIA Acc.} & \textbf{F1} & \textbf{F1(P)} & \textbf{MIA AUC} & \textbf{AIA Acc.} & \textbf{F1} & \textbf{F1(P)} & \textbf{MIA AUC} & \textbf{AIA Acc.}\\
\hline
Sentiment Analysis & 0.84 & 0.84 & 0.62 & 0.57 & 0.86 & 0.85 & 0.66 & 0.61 & 0.33 & 0.33 & 0.55 & 0.50\\
Emotion Detection & 0.62 & 0.58 & 0.70 & 0.63 & 0.59 & 0.56 & 0.68 & 0.60 & 0.52 & 0.31 & 0.72 & 0.65\\
Offensive Language Identification & 0.84 & 0.82 & 0.74 & 0.69 & 0.86 & 0.83 & 0.76 & 0.72 & 0.79 & 0.72 & 0.71 & 0.65\\
Code-Mixing & 0.63 & 0.57 & 0.69 & 0.66 & 0.59 & 0.54 & 0.68 & 0.64 & 0.63 & 0.60 & 0.67 & 0.61\\
Native Language Identification & 0.58 & 0.35 & 0.81 & 0.75 & 0.49 & 0.27 & 0.79 & 0.73 & 0.19 & 0.12 & 0.76 & 0.70\\
Dialect Identification & 0.64 & 0.60 & 0.73 & 0.68 & 0.55 & 0.42 & 0.77 & 0.72 & 0.41 & 0.33 & 0.75 & 0.70\\
\hline
\end{tabular}%
}

\caption{Finetuning (FT) Results. F1: F1 score of direct finetuning; F1(P): F1 score of privacy-preserved finetuning.}
\label{tab:results}
\end{table*}

\textbf{Bias, Fairness, and Discrimination Risks:} Ensuring fairness in NLP remains challenging, as large language models (LLMs) inherit biases from training data, resulting in unequal outcomes across gender, race, dialect, and region. For instance,~\citet{fleisig-etal-2024-linguistic} highlight dialect identification challenges, showing how opaque proprietary model training can lead to performance inequities across linguistic groups. Similarly,~\citet{tang2024gendercare} expose gender disparities in language model outputs, illustrating systemic biases from imbalanced datasets. In parallel,~\citet{beytia2022visual} reveal how gender biases embedded in large-scale resources like Wikipedia propagate into NLP systems, influencing their behavior in consequential ways. These three studies collectively underscore the urgent need for fairness-aware training and critical evaluation of model outputs. 

\textbf{Regulatory Compliance and Ethics:} Regulatory Compliance and Ethics are essential for responsible NLP research, yet only $8\%$ of the reviewed papers effectively address these issues. For instance,~\citet{xiao2018mobile} and~\citet{teodorescu2023language} could enhance compliance through clearer anonymization, while~\citet{raihan2023sentmix,raihan2024emomix} stress standardized procedures for code-mixing data collection. Likewise,~\citet{huang2015improved} emphasize documentation in dialect processing, and~\citet{fleisig-etal-2024-linguistic} with~\citet{hidayatullah2022systematic} highlight ethical risks in language discrimination and code-mixing. Furthermore,~\citet{jeong-etal-2022-kold} and~\citet{zampieri-etal-2024-federated} call for transparency in offensive language identification, while~\citet{jauhiainen2019automatic} and~\citet{mohamed2022artelingo} advocate ethical representation in multilingual datasets. Studies on code-switching, such as~\citet{garg2018code} and~\citet{nguyen2021automatic}, expose annotation and profiling risks, and~\citet{faisal-etal-2024-dialectbench} present a large-scale dataset requiring regulatory scrutiny. 

\subsection{Experimental Results}

\begin{table}[!htp]
\centering

\resizebox{\linewidth}{!}{
\begin{tabular}{l|ccc}
\hline
\textbf{Task} & \textbf{XLM-R} & \textbf{GPT-2} & \textbf{FLAN-T5} \\
\hline
Sentiment Analysis & \cellcolor{minor} Minor & \cellcolor{minor} Minor & \cellcolor{minor} Minor \\
Emotion Detection & \cellcolor{moderate} Moderate & \cellcolor{moderate} Moderate & \cellcolor{high} High \\
Offensive Language Identification & \cellcolor{minor} Minor & \cellcolor{minor} Minor & \cellcolor{moderate} Moderate \\
Code-Mixing & \cellcolor{moderate} Moderate & \cellcolor{moderate} Moderate & \cellcolor{minor} Minor \\
Native Language Identification & \cellcolor{high} High & \cellcolor{high} High & \cellcolor{high} High \\
Dialect Identification & \cellcolor{moderate} Moderate & \cellcolor{high} High & \cellcolor{high} High \\
\hline
\end{tabular}
}

\caption{\small{Performance Trade-off with Transformer Models Across NLP Tasks. \textcolor{minor}{\large\textbullet} Minor = $\leq$ 5\% 
\textcolor{moderate}{\large\textbullet} Moderate = 5\% $<$ F1 $\leq$ 10\% 
\textcolor{high}{\large\textbullet} High $\geq$ 10\%}}
\label{tab:discussion}
\end{table}

\begin{table*}[ht]
\centering
\resizebox{\linewidth}{!}{%
\renewcommand{\arraystretch}{1}
\rowcolors{2}{gray!5}{white}
\begin{tabular}{l|cccccc}
\hline
\textbf{NLP Task} & 
\textbf{\shortstack[c]{Data\\Collection}} & 
\textbf{\shortstack[c]{Anonymization}} & 
\textbf{\shortstack[c]{Visibility \&\\Profiling}} & 
\textbf{\shortstack[c]{Bias \&\\Fairness}} & 
\textbf{\shortstack[c]{Computational\\Privacy}} & 
\textbf{\shortstack[c]{Compliance \&\\Ethics}} \\
\hline
Sentiment Analysis      & \warn & \cmark & \xmark & \warn & \warn & \xmark \\
Emotion Detection       & \warn & \cmark & \cmark & \warn & \xmark & \xmark \\
Offensive Language      & \cmark & \cmark & \cmark & \warn & \xmark & \xmark \\
Code-Mixed Processing   & \cmark & \warn & \cmark & \warn & \xmark & \xmark \\
Native Language Identification      & \cmark & \cmark & \warn & \warn & \warn & \xmark \\
Dialect Identification  & \cmark & \cmark & \cmark & \cmark & \warn & \xmark \\
\hline

\end{tabular}
}
\vspace{1mm}

\cmark = Well-covered in literature, 
\warn = Partially addressed, 
\xmark = Largely missing.

\caption{Literature Gaps Across NLP Tasks Based on NLP-PRISM Dimensions}
\label{tab:appendix-lit-gaps}
\end{table*}

In line with the privacy evaluation outlined in NLP-PRISM, we conducted experiments to assess the impact of privacy-aware methodologies across six NLP tasks (Table~\ref{tab:results}). We used the binary-labeled dataset Sentiment140 for Sentiment Analysis~\cite{go2009twitter}, MELD~\cite{poria2019meld} with seven emotion classes for Emotion detection, and OLID~\cite{zampieri2019predicting} for Offensive Language Identification. Code-Mixing experiments employed the Malayalam-English dataset from Dravidian CodeMix at FIRE 2020~\cite{chakravarthi2020sentiment}, while TOEFL11~\cite{blanchard2013toefl11} with eleven L1s was used for Native Language Identification. Dialect Identification experiment combined English and Spanish dialect corpora from the VarDial 2024 shared task~\cite{chifu-etal-2024-vardial}, yielding nine dialect labels. Additional experimental results are summarized in Appendix Table~\ref{tab:results1}.

We fine-tuned XLM-R (encoder), GPT-2 (decoder), and FLAN-T5 (encoder–decoder) on task-specific datasets, integrating anonymization and noise addition. Following NER~\cite{sharma2022named}, entities such as Person (PER), Organization (ORG), Location (LOC), and Geo-Political Entities (GPE) were masked to reduce re-identification risks. Noise was introduced through 5\% text perturbations (character swaps, insertions, deletions, vowel-to-random-consonant substitutions) to mitigate inference attacks. These methods reduced data visibility and profiling risks: PII masking limited exposure, while noise reduced re-identification from model weights, addressing computational privacy. The anonymized, noise-resilient data further supported fairness and ethical AI compliance. Optimal hyperparameters were: batch size 8, learning rate 1e-5, epochs 3, and noise 0.05, with experiments run on a 40 GB institutional GPU cluster.

The privacy-utility trade-off varied across tasks and architectures (Table~\ref{tab:discussion}). Sentiment Analysis remained stable with minor degradation in GPT-2, while Emotion Detection showed greater drops, especially for FLAN-T5. Offensive Language Identification was least affected due to strong lexical cues. Code-Mixing tasks showed small declines, with FLAN-T5 displaying higher resilience. Native Language Identification suffered the greatest loss, particularly in GPT-2 and FLAN-T5, whereas XLM-R's multilingual pretraining ensured robustness; it also performed best in Dialect Identification. Overall, encoder-only models like XLM-R demonstrated superior stability under privacy constraints. Finally, membership inference attacks (MIA) (Table~\ref{tab:results}) were performed using shadow models to test adversarial distinguishability, while attribute inference attacks (AIA) (Table~\ref{tab:results}) were evaluated via probing classifiers predicting protected attributes (e.g., dialect, gender) from model representations.

\section{Discussion}
Social media NLP presents multifaceted risks arising from the intersection of language, identity, and algorithmic inference. Our survey addresses the core research question by identifying privacy vulnerabilities across six NLP tasks. However, we found that \textit{only 2\%–9\% of the papers across different tasks explicitly discuss privacy preservation} in our comprehensive analysis conducted within the NLP-PRISM framework (Table~\ref{tab:appendix-lit-gaps}).

Sentiment analysis and emotion detection use behavioral data that infer personal attributes like mental health or political stance. These risks heighten in sentiment (Yelp, Sentiment140, SentMix-3L) and emotion datasets (MELD, EmoMix-3L). \textit{About 63\% of sentiment analysis and 62\% of emotion detection papers haven't addressed privacy issues such as data collection, anonymization, or profiling.} Offensive language identification faces fairness and privacy risks. Datasets like OLID and SOLID yield high-performing models but embed biases harming minority dialects. Transformers frequently misclassify dialect-rich text, showing representativeness gaps; \textit{73\% of studies exhibited such issues leading to user profiling and computational privacy risks.} Code-mixed text processing introduces distinct privacy threats due to multilingual blending and transliteration, which can expose identity across languages. Datasets such as Hinglish, Tamlish, and Tenglish lack standardized anonymization, causing inconsistent protection. 

We identified~\textit{data anonymization, visibility, and compliance risks in 80\% of code-mixed studies}. Native language and dialect identification tasks exhibited a high re-identification potential through linguistic fingerprinting.~\textit{In $55\%$ of cases, even anonymized text was traceable via stylistic and syntactic cues}, with datasets such as Reddit-L2, AfriDial, and TwitterAAE found to be particularly vulnerable. Dialect datasets further posed geo-linguistic profiling risks, potentially revealing users' locations or social status; such issues were observed in~\textit{$69\%$ of dialect identification papers}. Moreover,~\textit{$40\%$ of the analyzed papers} cautioned that unanonymized competition datasets and LLMs increase the likelihood of identity disclosure.

While analyzing privacy risks, we also examined mitigation strategies (Appendix Table~\ref{tab:result7}) and found only $19$ papers addressing them. Key approaches include identity protection (e.g., named-entity masking, topic filtering)~\cite{teodorescu2023language,leonardelli2021agreeing,hidayatullah2022systematic}, model explainability tools (xAI, SHAP, LIME)~\cite{xiao2018mobile,mohamed2022artelingo}, and secured computation (SMPC, HE)~\cite{arango2024multifold,mehta2022social}. Data refinement methods such as phonetic normalization and aggregation~\cite{jeong-etal-2022-kold,sigurbergsson-derczynski-2020-offensive,berzak-etal-2017-predicting,ferragne2007automatic} further reduce re-identification risks. Beyond the six NLP tasks, NLP-PRISM can also extend to other social and domain-specific applications with privacy risks, enabling systematic analysis of vulnerabilities in areas like misinformation detection, stance classification, user profiling, and privacy-sensitive fields such as healthcare, education, and finance.

\section{Conclusion}
We conducted a comprehensive analysis of privacy risks in social media NLP, synthesizing evidence from $203$ studies across six core tasks. Using the \textit{NLP-PRISM} framework, we identified systemic vulnerabilities across data handling, feature exposure, computational processes, fairness, and regulatory compliance. The findings show that fewer than $10\%$ of studies address privacy in deployment contexts, and only a small subset incorporate formal threat modeling. Empirical evaluations with XLM-R, GPT-2, and FLAN-T5 indicate substantial information leakage (MIA AUC up to $0.81$, AIA accuracy up to $0.75$) and significant performance degradation (F1 reductions of up to $23\%$) under privacy-preserving fine-tuning. These results underscore the need for standardized privacy evaluation protocols, reproducible benchmarking, and regulation-aligned auditing to foster secure, fair, and socially responsible NLP systems.

\section*{Limitations}
In this work, we analyze peer-reviewed research on privacy risks in social media NLP. While peer-reviewed studies provide a reliable foundation for systematic analysis, we acknowledge that important research often appears on pre-print platforms and may not be covered within the scope of this work. Moreover, empirical validation in real-world applications and integration across NLP models remain ongoing challenges. Additionally, emerging generative AI models introduce new risks, including adversarial attacks and memorization vulnerabilities, which require further exploration. Moreover, it is found that, while applying privacy preserving techniques, native language and dialect identifications are the most susceptible in terms of performance due to the compromise of linguistic cues and it is difficult to balance this trade-off. Moving forward, we will refine privacy-aware techniques by analyzing on multiple datasets on downstream NLP tasks, improve fairness-aware learning, and develop adaptive methods that balance privacy with model utility. We will also implement explainable privacy indicators and real-time user controls to enhance transparency and user trust, ensuring more robust and privacy-aware NLP solutions.

\section*{Acknowledgment}
We would like to acknowledge the Data Agency and Security (DAS) Lab at George Mason University. The opinions expressed
in this work are solely those of the authors.

\bibliography{EACL}

@String{Computing = "Computing" }

@String{Computer = "{IEEE} Computer" }

@String{Academic = "Academic Press" }

@String{Springer = "Springer-Verlag" }

@inproceedings{lohar2021irish,
  title={Irish attitudes toward COVID tracker app \& privacy: sentiment analysis on Twitter and survey data},
  author={Lohar, Pintu and Xie, Guodong and Bendechache, Malika and Brennan, Rob and Celeste, Edoardo and Trestian, Ramona and Tal, Irina},
  booktitle={Proceedings of ARES},
  year={2021}
}

@article{xiao2018mobile,
  title={Mobile personalized service recommender model based on sentiment analysis and privacy concern},
  author={Xiao, Liang and Guo, Fei-Peng and Lu, Qi-Bei},
  journal={Mobile Information Systems},
  volume={2018},
  year={2018},
  publisher={Wiley Online Library}
}

@article{sharma2020role,
  title={Role of sentiment analysis in social media security and analytics},
  author={Sharma, Sanur and Jain, Anurag},
  journal={Wiley Interdisciplinary Reviews: Data Mining and Knowledge Discovery},
  volume={10},
  year={2020},
  publisher={Wiley Online Library}
}

@article{konate2018sentiment,
  title={Sentiment analysis of code-mixed Bambara-French social media text using deep learning techniques},
  author={Konate, Arouna and Du, Ruiying},
  journal={Wuhan University Journal of Natural Sciences},
  volume={23},
  year={2018},
  publisher={Springer}
}

@inproceedings{raihan2023sentmix,
  title={Sent{M}ix-3{L}: A novel code-mixed test dataset in bangla-english-hindi for sentiment analysis},
  author={Raihan, Md Nishat and Goswami, Dhiman and Mahmud, Antara and Anastasopoulos, Antonios and Zampieri, Marcos},
  booktitle={Proceedings of SEALP},
  year={2023}
}

@article{adhikari2025natural,
  title={Natural Language Processing of Privacy Policies: A Survey},
  author={Adhikari, Andrick and Das, Sanchari and Dewri, Rinku},
  journal={arXiv preprint arXiv:2501.10319},
  year={2025}
}

@inproceedings{gupta2024really,
  title={" I really just leaned on my community for support": Barriers, Challenges, and Coping Mechanisms Used by Survivors of $\{$Technology-Facilitated$\}$ Abuse to Seek Social Support},
  author={Gupta, Naman and Walsh, Kate and Das, Sanchari and Chatterjee, Rahul},
  booktitle={33rd USENIX Security Symposium (USENIX Security 24)},
  pages={4981--4998},
  year={2024}
}

@inproceedings{naveen2026privacy,
  title={Privacy Discourse and Emotional Dynamics in Mental Health Information Interaction on Reddit},
  author={Naveen Kumar, Jai Kruthunz and Surani, Aishwarya and Singh, Harkirat and Das, Sanchari},
  booktitle={Proceedings of the 2026 ACM SIGIR Conference on Human Information Interaction and Retrieval (CHIIR), Seattle, WA, USA},
  year={2026}
}

@inproceedings{zampieri-etal-2024-federated,
    title = "A Federated Learning Approach to Privacy Preserving Offensive Language Identification",
    author = "Zampieri, Marcos  and
      Premasiri, Damith  and
      Ranasinghe, Tharindu",
    editor = "Kumar, Ritesh  and
      Ojha, Atul Kr.  and
      Malmasi, Shervin  and
      Chakravarthi, Bharathi Raja  and
      Lahiri, Bornini  and
      Singh, Siddharth  and
      Ratan, Shyam",
    booktitle = "Proceedings of LREC-COLING",
    year = "2024",
}

@inproceedings{leonardelli2021agreeing,
  title={Agreeing to Disagree: Annotating Offensive Language Datasets with Annotators’ Disagreement},
  author={Leonardelli, Elisa and Menini, Stefano and Aprosio, Alessio Palmero and Guerini, Marco and Tonelli, Sara},
  booktitle={Proceedings of EMNLP},
  year={2021}
}

@inproceedings{davidson2017automated,
  title={Automated hate speech detection and the problem of offensive language},
  author={Davidson, Thomas and Warmsley, Dana and Macy, Michael and Weber, Ingmar},
  booktitle={Proceedings of ICWSM},
  volume={11},
  year={2017}
}

@inproceedings{deng-etal-2022-cold,
    title = "{COLD}: A Benchmark for {C}hinese Offensive Language Detection",
    author = "Deng, Jiawen  and
      Zhou, Jingyan  and
      Sun, Hao  and
      Zheng, Chujie  and
      Mi, Fei  and
      Meng, Helen  and
      Huang, Minlie",
    booktitle = "Proceedings of EMNLP",
    year = "2022",
}

@inproceedings{rizwan2020hate,
  title={Hate-speech and offensive language detection in roman Urdu},
  author={Rizwan, Hammad and Shakeel, Muhammad Haroon and Karim, Asim},
  booktitle={Proceedings of EMNLP},
  year={2020}
}

@inproceedings{jeong-etal-2022-kold,
    title = "{KOLD}: {K}orean Offensive Language Dataset",
    author = "Jeong, Younghoon  and
      Oh, Juhyun  and
      Lee, Jongwon  and
      Ahn, Jaimeen  and
      Moon, Jihyung  and
      Park, Sungjoon  and
      Oh, Alice",
    booktitle = "Proceedings of EMNLP",
    year = "2022",
}

@inproceedings{arango2024multifold,
  title={MultiFOLD: Multi-source Domain Adaption for Offensive Language Detection},
  author={Arango, Aym{\'e} and Kaghazgaran, Parisa and Sarwar, Sheikh Muhammad and Murdock, Vanessa and Lee, Cj},
  booktitle={Proceedings of ICWSM},
  volume={18},
  pages={86--99},
  year={2024}
}

@inproceedings{sigurbergsson-derczynski-2020-offensive,
    title = "Offensive Language and Hate Speech Detection for {D}anish",
    author = "Sigurbergsson, Gudbjartur Ingi  and
      Derczynski, Leon",
    booktitle = "Proceedings of LREC",
    year = "2020",
}

@inproceedings{pitenis-etal-2020-offensive,
    title = "Offensive Language Identification in {G}reek",
    author = "Pitenis, Zesis  and
      Zampieri, Marcos  and
      Ranasinghe, Tharindu",
    booktitle = "Proceedings of LREC",
    year = "2020",
}

@inproceedings{goswami2023offmix,
  title={OffMix-3L: A novel code-mixed test dataset in bangla-english-hindi for offensive language identification},
  author={Goswami, Dhiman and Raihan, Md Nishat and Mahmud, Antara and Anastasopoulos, Antonios and Zampieri, Marcos},
  booktitle={Proceedings of SocialNLP},
  year={2023}
}

@article{trajano2024olid,
  title={Olid-br: offensive language identification dataset for brazilian portuguese},
  author={Trajano, Douglas and Bordini, Rafael H and Vieira, Renata},
  journal={Language Resources and Evaluation},
  volume={58},
  pages={1263--1289},
  year={2024},
  publisher={Springer}
}

@inproceedings{rosenthal2021solid,
  title={SOLID: A Large-Scale Semi-Supervised Dataset for Offensive Language Identification},
  author={Rosenthal, Sara and Atanasova, Pepa and Karadzhov, Georgi and Zampieri, Marcos and Nakov, Preslav},
  booktitle={Findings of ACL-IJCNLP},
  year={2021}
}

@inproceedings{morabito2024stop,
  title={STOP! Benchmarking Large Language Models with Sensitivity Testing on Offensive Progressions},
  author={Morabito, Robert and Madhusudan, Sangmitra and McDonald, Tyler and Emami, Ali},
  booktitle={Proceedings of EMNLP},
  year={2024}
}

@inproceedings{xiao-etal-2024-toxicloakcn,
    title = "{T}oxi{C}loak{CN}: Evaluating Robustness of Offensive Language Detection in {C}hinese with Cloaking Perturbations",
    author = "Xiao, Yunze  and
      Hu, Yujia  and
      Choo, Kenny Tsu Wei  and
      Lee, Roy Ka-Wei",
    editor = "Al-Onaizan, Yaser  and
      Bansal, Mohit  and
      Chen, Yun-Nung",
    booktitle = "Proceedings of EMNLP",
    year = "2024",
}

@inproceedings{mohamed2022artelingo,
  title={ArtELingo: A Million Emotion Annotations of WikiArt with Emphasis on Diversity over Language and Culture},
  author={Mohamed, Youssef and Abdelfattah, Mohamed and Alhuwaider, Shyma and Li, Feifan and Zhang, Xiangliang and Church, Kenneth and Elhoseiny, Mohamed},
  booktitle={Proceedings of EMNLP},
  year={2022}
}

@article{raihan2024emomix,
  title={EmoMix-3L: A Code-Mixed Dataset for Bangla-English-Hindi Emotion Detection},
  author={Raihan, Nishat and Goswami, Dhiman and Mahmud, Antara and Anastasopoulos, Antonios and Zampieri, Marcos},
  journal={LREC-COLING},
  pages={11},
  year={2024}
}

@inproceedings{canales2014emotion,
  title={Emotion detection from text: A survey},
  author={Canales, Lea and Mart{\'\i}nez-Barco, Patricio},
  booktitle={Proceedings of JISIC},
  pages={37--43},
  year={2014}
}

@inproceedings{teodorescu2023language,
  title={Language and Mental Health: Measures of Emotion Dynamics from Text as Linguistic Biosocial Markers},
  author={Teodorescu, Daniela and Cheng, Tiffany and Fyshe, Alona and Mohammad, Saif},
  booktitle={Proceedings of EMNLP},
  year={2023}
}

@inproceedings{gill2008language,
  title={The language of emotion in short blog texts},
  author={Gill, Alastair J and French, Robert M and Gergle, Darren and Oberlander, Jon},
  booktitle={Proceedings of CSCW},
  year={2008}
}

@inproceedings{sassi2024afridial,
  title={AfriDial: African Dialect Model based on Deep Learning for Sentiment Analysis},
  author={Sassi, Ameni and Tonga, Junior and Poaty, St{\'e}phanie and Steve, Sanon and Adjid, Djibrine Idriss Abakar and Cherif, Moukhtar and Ouarda, Wael},
  booktitle={Proceedings of IWCMC},
  year={2024},
}

@inproceedings{malmasi2015arabic,
  title={Arabic dialect identification using a parallel multidialectal corpus},
  author={Malmasi, Shervin and Refaee, Eshrag and Dras, Mark},
  booktitle={Proceedings of PACLING},
  year={2015},
}

@inproceedings{berrimi2020arabic,
  title={Arabic dialects identification: North African dialects case study},
  author={Berrimi, Mohamed and Moussaoui, Abdelouahab and Oussalah, Mourad and Saidi, Mohamed},
  booktitle={Proceedings of IAM},
  year={2020},
}

@inproceedings{zissman1996automatic,
  title={Automatic dialect identification of extemporaneous conversational, Latin American Spanish speech},
  author={Zissman, Marc A and Gleason, Terry P and Rekart, Deborah M and Losiewicz, Beth L},
  booktitle={Proceedings of ICASSP},
  year={1996},
}

@article{ferragne2007automatic,
  title={Automatic dialect identification: A study of British English},
  author={Ferragne, Emmanuel and Pellegrino, Fran{\c{c}}ois},
  journal={Speaker Classification II: Selected Projects},
  year={2007},
  publisher={Springer}
}

@inproceedings{jorgensen2015challenges,
  title={Challenges of studying and processing dialects in social media},
  author={J{\o}rgensen, Anna and Hovy, Dirk and S{\o}gaard, Anders},
  booktitle={Proceedings of WNUT},
  year={2015}
}

@article{mahmud2023cyberbullying,
  title={Cyberbullying detection for low-resource languages and dialects: Review of the state of the art},
  author={Mahmud, Tanjim and Ptaszynski, Michal and Eronen, Juuso and Masui, Fumito},
  journal={Information Processing \& Management},
  volume={60},
  year={2023},
  publisher={Elsevier}
}

@inproceedings{blodgett2016demographic,
  title={Demographic Dialectal Variation in Social Media: A Case Study of African-American English},
  author={Blodgett, Su Lin and Green, Lisa and O’Connor, Brendan},
  booktitle={Proceedings of EMNLP},
  year={2016}
}

@inproceedings{faisal-etal-2024-dialectbench,
    title = "{DIALECTBENCH}: An {NLP} Benchmark for Dialects, Varieties, and Closely-Related Languages",
    author = "Faisal, Fahim  and
      Ahia, Orevaoghene  and
      Srivastava, Aarohi  and
      Ahuja, Kabir  and
      Chiang, David  and
      Tsvetkov, Yulia  and
      Anastasopoulos, Antonios",
    booktitle = "Proceedings of ACL",
    year = "2024",
}

@inproceedings{huang2015improved,
  title={Improved Arabic dialect classification with social media data},
  author={Huang, Fei},
  booktitle={Proceedings of EMNLP},
  year={2015}
}

@inproceedings{fleisig-etal-2024-linguistic,
    title = "Linguistic Bias in {C}hat{GPT}: Language Models Reinforce Dialect Discrimination",
    author = "Fleisig, Eve  and
      Smith, Genevieve  and
      Bossi, Madeline  and
      Rustagi, Ishita  and
      Yin, Xavier  and
      Klein, Dan",
    booktitle = "Proceedings of EMNLP",
    year = "2024",
}

@inproceedings{barot2024tonecheck,
  title={ToneCheck: Unveiling the Impact of Dialects in Privacy Policy},
  author={Barot, Jay and Allami, Ali and Yin, Ming and Lin, Dan},
  booktitle={Proceedings of SACMAT},
  year={2024}
}

@inproceedings{bierner2001alternative,
  title={Alternative phrases and natural language information retrieval},
  author={Bierner, Gann},
  booktitle={Proceedings of ACL},
  year={2001}
}

@article{nguyen2021automatic,
  title={Automatic language identification in code-switched Hindi-English social media text},
  author={Nguyen, Li and Bryant, Christopher and Kidwai, Sana and Biberauer, Theresa},
  journal={Journal of Open Humanities Data},
  volume={7},
  year={2021}
}

@article{jauhiainen2019automatic,
  title={Automatic language identification in texts: A survey},
  author={Jauhiainen, Tommi and Lui, Marco and Zampieri, Marcos and Baldwin, Timothy and Lind{\'e}n, Krister},
  journal={Journal of Artificial Intelligence Research},
  volume={65},
  year={2019}
}

@inproceedings{staicu2023bilingual,
  title={Bilingual problems: Studying the security risks incurred by native extensions in scripting languages},
  author={Staicu, Cristian-Alexandru and Rahaman, Sazzadur and Kiss, {\'A}gnes and Backes, Michael},
  booktitle={Proceedings of USENIX},
  year={2023}
}

@inproceedings{aoyama2024modeling,
  title={Modeling Nonnative Sentence Processing with L2 Language Models},
  author={Aoyama, Tatsuya and Schneider, Nathan},
  booktitle={Proceedings of EMNLP},
  year={2024}
}

@inproceedings{goldin2018native,
  title={Native language identification with user generated content},
  author={Goldin, Gili and Rabinovich, Ella and Wintner, Shuly},
  booktitle={Proceedings of EMNLP},
  year={2018}
}

@inproceedings{berzak-etal-2017-predicting,
    title = "Predicting Native Language from Gaze",
    author = "Berzak, Yevgeni  and
      Nakamura, Chie  and
      Flynn, Suzanne  and
      Katz, Boris",
    booktitle = "Proceedings of ACL",
    year = "2017",
}

@inproceedings{buschek2021impact,
  title={The impact of multiple parallel phrase suggestions on email input and composition behaviour of native and non-native english writers},
  author={Buschek, Daniel and Z{\"u}rn, Martin and Eiband, Malin},
  booktitle={Proceedings of CHI},
  year={2021}
}

@inproceedings{lamabam2016language,
  title={A language identification system for code-mixed English-Manipuri Social Media text},
  author={Lamabam, Priyadarshini and Chakma, Kunal},
  booktitle={Proceedings of ICETECH},
  year={2016},
}

@inproceedings{dougruoz2021survey,
  title={A Survey of Code-switching: Linguistic and Social Perspectives for Language Technologies},
  author={Do{\u{g}}ru{\"o}z, A Seza and Sitaram, Sunayana and Bullock, Barbara and Toribio, Almeida Jacqueline},
  booktitle={Proceedings of ACL - IJCNLP},
  year={2021}
}

@article{hidayatullah2022systematic,
  title={A systematic review on language identification of code-mixed text: techniques, data availability, challenges, and framework development},
  author={Hidayatullah, Ahmad Fathan and Qazi, Atika and Lai, Daphne Teck Ching and Apong, Rosyzie Anna},
  journal={IEEE access},
  volume={10},
  year={2022},
  publisher={IEEE}
}

@article{nabila2022analysis,
  title={An Analysis of Indonesian-Eglish Code Mixing Used in Social Media (TWITTER)},
  author={Nabila, Cindy and Idayani, Andi},
  journal={J-SHMIC: Journal of English for Academic},
  volume={9},
  year={2022}
}

@inproceedings{lakshmi2017automatic,
  title={An automatic language identification system for code-mixed English-Kannada social media text},
  author={Lakshmi, BS Sowmya and Shambhavi, BR},
  booktitle={Proceedings of CSITSS},
  year={2017},
}

@article{tabe2023code,
  title={Code-Mixing and Code-Switching in Cameroon Social Media},
  author={Tabe, Camilla Arundie},
  journal={International Journal of Linguistics and Translation Studies},
  volume={4},
  year={2023}
}

@inproceedings{garg2018code,
  title={Code-switched Language Models Using Dual RNNs and Same-Source Pretraining},
  author={Garg, Saurabh and Parekh, Tanmay and Jyothi, Preethi},
  booktitle={Proceedings of EMNLP},
  year={2018}
}

@inproceedings{shanmugalingam2019language,
  title={Language identification at word level in Sinhala-English code-mixed social media text},
  author={Shanmugalingam, Kasthuri and Sumathipala, Sagara},
  booktitle={Proceedings of SCSE},
  year={2019},
}

@article{sabty2021language,
  title={Language identification of intra-word code-switching for Arabic--English},
  author={Sabty, Caroline and Mesabah, Islam and {\c{C}}etino{\u{g}}lu, {\"O}zlem and Abdennadher, Slim},
  journal={Array},
  volume={12},
  year={2021},
  publisher={Elsevier}
}

@inproceedings{zhang2023multilingual,
  title={Multilingual Large Language Models Are Not (Yet) Code-Switchers},
  author={Zhang, Ruochen and Cahyawijaya, Samuel and Cruz, Jan Christian Blaise and Winata, Genta and Aji, Alham},
  booktitle={Proceedings of EMNLP},
  year={2023}
}

@inproceedings{strengers2020adhering,
  title={Adhering, steering, and queering: Treatment of gender in natural language generation},
  author={Strengers, Yolande and Qu, Lizhen and Xu, Qiongkai and Knibbe, Jarrod},
  booktitle={Proceedings of CHI},
  year={2020}
}

@inproceedings{lukas2023analyzing,
  title={Analyzing leakage of personally identifiable information in language models},
  author={Lukas, Nils and Salem, Ahmed and Sim, Robert and Tople, Shruti and Wutschitz, Lukas and Zanella-B{\'e}guelin, Santiago},
  booktitle={Proceedings of S\&P},
  year={2023},
  organization={IEEE}
}

@inproceedings{hounsel2018automatically,
  title={Automatically Generating a Large,$\{$Culture-Specific$\}$ Blocklist for China},
  author={Hounsel, Austin and Mittal, Prateek and Feamster, Nick},
  booktitle={Proceedings of FOCI},
  year={2018}
}

@inproceedings{xu2024characterization,
  title={Characterization of Political Polarized Users Attacked by Language Toxicity on Twitter},
  author={Xu, Wentao},
  booktitle={Proceedings of CSCW},
  year={2024}
}

@article{dewani2023detection,
  title={Detection of cyberbullying patterns in low resource colloquial Roman Urdu microtext using natural language processing, machine learning, and ensemble techniques},
  author={Dewani, Amirita and Memon, Mohsin Ali and Bhatti, Sania and Sulaiman, Adel and Hamdi, Mohammed and Alshahrani, Hani and Alghamdi, Abdullah and Shaikh, Asadullah},
  journal={Applied Sciences},
  volume={13},
  year={2023},
  publisher={MDPI}
}

@inproceedings{tang2024gendercare,
  title={GenderCARE: A comprehensive framework for assessing and reducing gender bias in large language models},
  author={Tang, Kunsheng and Zhou, Wenbo and Zhang, Jie and Liu, Aishan and Deng, Gelei and Li, Shuai and Qi, Peigui and Zhang, Weiming and Zhang, Tianwei and Yu, Nenghai},
  booktitle={Proceedings of CCS},
  year={2024}
}

@inproceedings{stokes2023language,
  title={How language formality in security and privacy interfaces impacts intended compliance},
  author={Stokes, Jackson and August, Tal and Marver, Robert A and Czeskis, Alexei and Roesner, Franziska and Kohno, Tadayoshi and Reinecke, Katharina},
  booktitle={Proceedings of CHI},
  year={2023}
}

@inproceedings{vishwamitra2024moderating,
  title={Moderating new waves of online hate with chain-of-thought reasoning in large language models},
  author={Vishwamitra, Nishant and Guo, Keyan and Romit, Farhan Tajwar and Ondracek, Isabelle and Cheng, Long and Zhao, Ziming and Hu, Hongxin},
  booktitle={Proceedings of IEEE S\&P},
  year={2024},
}

@inproceedings{eleta2012multilingual,
  title={Multilingual use of twitter: social networks and language choice},
  author={Eleta, Irene},
  booktitle={Proceedings of CSCW},
  year={2012}
}

@inproceedings{reitmaier2022opportunities,
  title={Opportunities and challenges of automatic speech recognition systems for low-resource language speakers},
  author={Reitmaier, Thomas and Wallington, Electra and Kalarikalayil Raju, Dani and Klejch, Ondrej and Pearson, Jennifer and Jones, Matt and Bell, Peter and Robinson, Simon},
  booktitle={Proceedings of CHI},
  year={2022}
}

@inproceedings{beytia2022visual,
  title={Visual gender biases in wikipedia: A systematic evaluation across the ten most spoken languages},
  author={Beyt{\'\i}a, Pablo and Agarwal, Pushkal and Redi, Miriam and Singh, Vivek K},
  booktitle={Proceedings of ICWSM},
  year={2022}
}

@inproceedings{he2024you,
  title={You only prompt once: On the capabilities of prompt learning on large language models to tackle toxic content},
  author={He, Xinlei and Zannettou, Savvas and Shen, Yun and Zhang, Yang},
  booktitle={Proceedings of IEEE S\&P},
  year={2024},
}

@article{mcinnes2018preferred,
  title={Preferred reporting items for a systematic review and meta-analysis of diagnostic test accuracy studies: the PRISMA-DTA statement},
  author={McInnes, Matthew DF and Moher, David and Thombs, Brett D and McGrath, Trevor A and Bossuyt, Patrick M and Clifford, Tammy and Cohen, J{\'e}r{\'e}mie F and Deeks, Jonathan J and Gatsonis, Constantine and Hooft, Lotty and others},
  journal={Jama},
  volume={319},
  pages={388--396},
  year={2018},
  publisher={American Medical Association}
}

@article{moher2010preferred,
  title={Preferred reporting items for systematic reviews and meta-analyses: the PRISMA statement},
  author={Moher, David and Liberati, Alessandro and Tetzlaff, Jennifer and Altman, Douglas G and Prisma Group and others},
  journal={International journal of surgery},
  volume={8},
  pages={336--341},
  year={2010},
  publisher={Elsevier}
}

@article{ortiz2023implications,
  title={Implications of Emotion Recognition Technologies: Balancing Privacy and Public Safety},
  author={Ortiz-Clavijo, Luis Felipe and Gallego-Duque, Carlos Juli{\'a}n and David-Diaz, Juan Camilo and Ortiz-Zamora, Andr{\'e}s Felipe},
  journal={IEEE Technology and Society Magazine},
  volume={42},
  pages={69--75},
  year={2023},
  publisher={IEEE}
}

@inproceedings{malmasi2017report,
  title={A report on the 2017 native language identification shared task},
  author={Malmasi, Shervin and Evanini, Keelan and Cahill, Aoife and Tetreault, Joel and Pugh, Robert and Hamill, Christopher and Napolitano, Diane and Qian, Yao},
  booktitle={Proceedings of BEA},
  year={2017}
}

@article{zhang2023native,
  title={Native Language Identification with Large Language Models},
  author={Zhang, Wei and Salle, Alexandre},
  journal={arXiv preprint arXiv:2312.07819},
  year={2023}
}

@article{wang2017protecting,
  title={Protecting personal trajectories of social media users through differential privacy},
  author={Wang, Shuo and Sinnott, Richard O},
  journal={Computers \& Security},
  volume={67},
  pages={142--163},
  year={2017},
  publisher={Elsevier}
}

@article{singh2017importance,
  title={Importance and challenges of social media text},
  author={Singh, Shailendra Kumar and Manoj, KS},
  journal={International Journal of Advanced Research in Computer Science},
  volume={8},
  pages={831--834},
  year={2017}
}

@incollection{luca2015user,
  title={User-generated content and social media},
  author={Luca, Michael},
  booktitle={Handbook of media Economics},
  volume={1},
  pages={563--592},
  year={2015},
  publisher={Elsevier}
}

@article{beigi2020survey,
  title={A survey on privacy in social media: Identification, mitigation, and applications},
  author={Beigi, Ghazaleh and Liu, Huan},
  journal={ACM Transactions on Data Science},
  volume={1},
  pages={1--38},
  year={2020},
  publisher={ACM New York, NY, USA}
}

@article{das2024security,
  title={Security and privacy challenges of large language models: A survey},
  author={Das, Badhan Chandra and Amini, M Hadi and Wu, Yanzhao},
  journal={ACM Computing Surveys},
  year={2024},
  publisher={ACM New York, NY}
}

@article{mistry2024federated,
  title={Federated Learning-Based Architecture for Personalized Next Emoji Prediction for Social Media Comments},
  author={Mistry, Durjoy and Plabon, Jayonto Dutta and Diba, Bidita Sarkar and Mukta, Saddam and Mridha, MF},
  journal={IEEE Access},
  year={2024},
  publisher={IEEE}
}

@phdthesis{malmasi2016native,
  title={Native language identification: explorations and applications},
  author={Malmasi, Shervin and others},
  year={2016},
  school={Macquarie University, Faculty of Science and Engineering, Department of~…}
}

@article{meier2024llm,
  title={LLM-Aided Social Media Influence Operations},
  author={Meier, Raphael},
  journal={Large Language Models in Cybersecurity: Threats, Exposure and Mitigation},
  pages={105--112},
  year={2024},
  publisher={Springer Nature Switzerland Cham}
}

@inproceedings{gharehchopogh2011analysis,
  title={Analysis and evaluation of unstructured data: text mining versus natural language processing},
  author={Gharehchopogh, Farhad Soleimanian and Khalifelu, Zeinab Abbasi},
  booktitle={Proceedings of AICT},
  year={2011},
  organization={IEEE}
}

@article{blanchard2013toefl11,
  title={TOEFL11: A Corpus of Non-native English},
  author={Blanchard, D},
  journal={Educational Testing Service},
  year={2013}
}

@inproceedings{tetreault2013report,
  title={A report on the first native language identification shared task},
  author={Tetreault, Joel and Blanchard, Daniel and Cahill, Aoife},
  booktitle={Proceedings of BEA},
  year={2013}
}

@inproceedings{anand2017overview,
  title={Overview of the INLI PAN at FIRE-2017 track on Indian native language identification},
  author={Anand Kumar, M and Barathi Ganesh, HB and Singh, Shivkaran and Soman, KP and Rosso, Paolo},
  booktitle={CEUR workshop proceedings},
  year={2017}
}

@article{soman2018overview,
  title={Overview of the second shared task on Indian native language identification (INLI)},
  journal={CEUR Workshop},
  author={Soman, KP},
  year={2018}
}

@inproceedings{steinbakken2020native,
  title={Native-language identification with attention},
  author={Steinbakken, Stian and Gamb{\"a}ck, Bj{\"o}rn},
  booktitle={Proceedings of ICON},
  year={2020}
}

@inproceedings{koppel2005determining,
  title={Determining an author's native language by mining a text for errors},
  author={Koppel, Moshe and Schler, Jonathan and Zigdon, Kfir},
  booktitle={Proceedings of ACM SIGKDD},
  year={2005}
}

@inproceedings{wong2009contrastive,
  title={Contrastive analysis and native language identification},
  author={Wong, Sze-Meng Jojo and Dras, Mark},
  booktitle={Proceedings of ALTA},
  year={2009}
}

@inproceedings{mechti2016empirical,
  title={An empirical method using features combination for Arabic native language identification},
  author={Mechti, Seifeddine and Abbassi, Ayoub and Belguith, Lamia Hadrich and Faiz, Rim},
  booktitle={Proceedings of AICCSA},
  year={2016},
}

@inproceedings{gebre2013improving,
  title={Improving native language identification with tf-idf weighting},
  author={Gebre, Binyam Gebrekidan and Zampieri, Marcos and Wittenburg, Peter and Heskes, Tom},
  booktitle={Proceedings of BEA},
  year={2013}
}

@inproceedings{jarvis2013maximizing,
  title={Maximizing classification accuracy in native language identification},
  author={Jarvis, Scott and Bestgen, Yves and Pepper, Steve},
  booktitle={Proceedings of BEA},
  year={2013}
}

@inproceedings{di2020good,
  title={How good are humans at Native Language Identification? A case study on Italian L2 writings},
  author={Di Nuovo, Elisa and Bosco, Cristina and Corino, Elisa and others},
  booktitle={Proceedings of CLiC-it},
  year={2020},
}

@inproceedings{zampieri2017native,
  title={Native Language Identification on Text and Speech},
  author={Zampieri, Marcos and Ciobanu, Alina Maria and Dinu, Liviu P},
  booktitle={Proceedings of BEA},
  year={2017}
}

@inproceedings{raihan2023offensive,
  title={Offensive Language Identification in Transliterated and Code-Mixed Bangla},
  author={Raihan, Md Nishat and Tanmoy, Umma Hani and Islam, Anika Binte and North, Kai and Ranasinghe, Tharindu and Anastasopoulos, Antonios and Zampieri, Marcos},
  booktitle={Proceedings of BLP},
  year={2023}
}

@inproceedings{zampieri2019predicting,
  title={Predicting the Type and Target of Offensive Posts in Social Media},
  author={Zampieri, Marcos and Malmasi, Shervin and Nakov, Preslav and Rosenthal, Sara and Farra, Noura and Kumar, Ritesh},
  booktitle={Proceedings of NAACL},
  year={2019}
}

@inproceedings{samardzic2016archimob,
  title={Archimob-a corpus of spoken swiss german},
  author={Samardzic, Tanja and Scherrer, Yves and Glaser, Elvira},
  booktitle={Proceedings of LREC},
  year={2016}
}

@inproceedings{2022-findings-vardial,
 title = "Findings of the {V}ar{D}ial Evaluation Campaign 2022",
 author = "Aepli, No{\"e}mi  and
   Anastasopoulos, Antonios  and
   Chifu, Adrian  and
   Domingues, William  and
   Faisal, Fahim  and
   G{\u{a}}man, Mihaela  and
   Ionescu, Radu Tudor  and
   Scherrer, Yves",
 booktitle = "Proceedings of ICCL (VarDial)",
 year = "2022",
}

@article{chandu2018code,
  title={Code-Mixed Question Answering Challenge: Crowd-sourcing Data and Techniques},
  author={Chandu, Khyathi Raghavi and Loginova, Ekaterina and Gupta, Vishal and van Genabith, Josef and {\"u}nter Neuman, G and Chinnakotla, Manoj and Nyberg, Eric and Black, Alan},
  journal={ACL 2018},
  pages={29},
  year={2018}
}

@inproceedings{poria2019meld,
  title={MELD: A Multimodal Multi-Party Dataset for Emotion Recognition in Conversations},
  author={Poria, Soujanya and Hazarika, Devamanyu and Majumder, Navonil and Naik, Gautam and Cambria, Erik and Mihalcea, Rada},
  booktitle={Proceedings of ACL},
  year={2019}
}

@inproceedings{nojavanasghari2016emoreact,
  title={Emoreact: a multimodal approach and dataset for recognizing emotional responses in children},
  author={Nojavanasghari, Behnaz and Baltru{\v{s}}aitis, Tadas and Hughes, Charles E and Morency, Louis-Philippe},
  booktitle={Proceedings of ICMI},
  year={2016}
}

@article{go2009twitter,
  title={Twitter sentiment classification using distant supervision},
  author={Go, Alec and Bhayani, Richa and Huang, Lei},
  journal={CS224N project report, Stanford},
  volume={1},
  pages={2009},
  year={2009}
}

@article{zhang2015character,
  title={Character-level convolutional networks for text classification},
  author={Zhang, Xiang and Zhao, Junbo and LeCun, Yann},
  journal={Advances in neural information processing systems},
  volume={28},
  year={2015}
}

@inproceedings{ghosh2023annihilate,
  title={Annihilate Hates (Task 4 HASOC 2023): Hate Speech Detection in Assamese Bengali and Bodo languages.},
  author={Ghosh, Koyel and Senapati, Apurbalal and Pal, Aditya Shankar},
  booktitle={FIRE (Working Notes)},
  year={2023}
}

@article{ranasinghe2022overview,
  title={Overview of the HASOC Subtrack at FIRE 2022: Offensive Language Identification in Marathi},
  author={Ranasinghe, Tharindu and North, Kai and Premasiri, Damith and Zampieri, Marcos},
journal={CEUR Workshop},
  booktitle={FIRE (Working Notes)},
  year={2022}
}

@inproceedings{mandl2019overview,
  title={Overview of the hasoc track at fire 2019: Hate speech and offensive content identification in indo-european languages},
  author={Mandl, Thomas and Modha, Sandip and Majumder, Prasenjit and Patel, Daksh and Dave, Mohana and Mandlia, Chintak and Patel, Aditya},
  booktitle={Proceedings of FIRE},
  year={2019}
}

@inproceedings{mandl2020overview,
  title={Overview of the hasoc track at fire 2020: Hate speech and offensive language identification in tamil, malayalam, hindi, english and german},
  author={Mandl, Thomas and Modha, Sandip and Kumar M, Anand and Chakravarthi, Bharathi Raja},
  booktitle={Proceedings of FIRE},
  year={2020}
}

@inproceedings{kumar-etal-2018-benchmarking,

title = "Benchmarking Aggression Identification in Social Media",
author = "Kumar, Ritesh and Ojha, Atul Kr. and Malmasi, Shervin and Zampieri, Marcos",
booktitle = "Proceedings of TRAC",
year = "2018",
}

@inproceedings{risch-etal-2021-overview,
    title = "Overview of the {G}erm{E}val 2021 Shared Task on the Identification of Toxic, Engaging, and Fact-Claiming Comments",
    author = "Risch, Julian  and
      Stoll, Anke  and
      Wilms, Lena  and
      Wiegand, Michael",
    booktitle = "Proceedings of GermEval",
    year = "2021",
}

@inproceedings{toxicspans-acl,
title={Findings of the Shared Task on {O}ffensive {S}pan {I}dentification in Code-Mixed {T}amil-{E}nglish Comments},
author = "Ravikiran, Manikandan and
Chakravarthi, Bharathi Raja and
Madasamy, Anand Kumar and
Sivanesan, Sangeetha and
Rajalakshmi, Ratnavel and
Thavareesan, Sajeetha and
Ponnusamy, Rahul and
Mahadevan, Shankar",
booktitle = "Proceedings of DravidianLangTech",
year = "2022",
}

@inproceedings{premjith-2024-eacl-hate-telugu,    
title = "Findings of the Shared Task on Hate and Offensive Language Detection in Telugu Codemixed Text (HOLD-Telugu)",    
author = "B, Premjth and Chakravarthi, Bharathi Raja and Kumaresan, Prasanna Kumar and Rajiakodi, Saranya and Karnati, Sai Prashanth and Mangamuru, Sai Rishith Reddy and Chandu, Janakiram",    
booktitle = "Proceedings of DravidianLangTech",       
year = "2024",
}

@inproceedings{ravikiran-etal-2022-findings,
    title = "Findings of the Shared Task on Offensive Span Identification from{C}ode-Mixed {T}amil-{E}nglish Comments",
    author = "Ravikiran, Manikandan  and
      Chakravarthi, Bharathi Raja  and
      Madasamy, Anand Kumar  and
      S, Sangeetha  and
      Rajalakshmi, Ratnavel  and
      Thavareesan, Sajeetha  and
      Ponnusamy, Rahul  and
      Mahadevan, Shankar",
    booktitle = "Proceedings of DravidianLangTech",
    year = "2022",
}

@inproceedings{chakravarthi-etal-2021-findings-shared,
    title = "Findings of the Shared Task on Offensive Language Identification in {T}amil, {M}alayalam, and {K}annada",
    author = "Chakravarthi, Bharathi Raja  and
      Priyadharshini, Ruba  and
      Jose, Navya  and
      Kumar M, Anand  and
      Mandl, Thomas  and
      Kumaresan, Prasanna Kumar  and
      Ponnusamy, Rahul  and
      R L, Hariharan  and
      McCrae, John P.  and
      Sherly, Elizabeth",
    booktitle = "Proceedings of DravidianLangTech",
    year = "2021",
}

@inProceedings{wang2024SemEval,
  author={Wang, Fanfan  and  Ma, Heqing  and  Xia, Rui  and  Yu, Jianfei  and  Cambria, Erik},
  title = "{S}em{E}val-2024 Task 3: Multimodal Emotion Cause Analysis in Conversations",
  booktitle = "Proceedings of SemEval",
  year = "2024",
}

@inproceedings{kumar-etal-2024-semeval,
    title = "{S}em{E}val 2024 - Task 10: Emotion Discovery and Reasoning its Flip in Conversation ({ED}i{R}e{F})",
    author = "Kumar, Shivani  and
      Akhtar, Md. Shad  and
      Cambria, Erik  and
      Chakraborty, Tanmoy",
    booktitle = "Proceedings of SemEval",
    year = "2024",
}

@inproceedings{muhammad-etal-2023-semeval,
    title = "{S}em{E}val-2023 Task 12: Sentiment Analysis for {A}frican Languages ({A}fri{S}enti-{S}em{E}val)",
    author = "Muhammad, Shamsuddeen Hassan  and
      Abdulmumin, Idris  and
      Yimam, Seid Muhie  and
      Adelani, David Ifeoluwa  and
      Ahmad, Ibrahim Said  and
      Ousidhoum, Nedjma  and
      Ayele, Abinew Ali  and
      Mohammad, Saif  and
      Beloucif, Meriem  and
      Ruder, Sebastian",
    booktitle = "Proceedings of SemEval",
    year = "2023",
}

@inproceedings{barnes-etal-2022-semeval,
    title = "{S}em{E}val 2022 Task 10: Structured Sentiment Analysis",
    author = "Barnes, Jeremy  and
      Oberlaender, Laura  and
      Troiano, Enrica  and
      Kutuzov, Andrey  and
      Buchmann, Jan  and
      Agerri, Rodrigo  and
      {\O}vrelid, Lilja  and
      Velldal, Erik",
    booktitle = "Proceedings of SemEval",
    year = "2022",
}

@inproceedings{patwa2020sentimix,
title={SemEval-2020 Task 9: Overview of Sentiment Analysis of Code-Mixed Tweets},
author={Patwa, Parth and
Aguilar, Gustavo and
Kar, Sudipta and
Pandey, Suraj and
PYKL, Srinivas and
Gamb{\"a}ck, Bj{\"o}rn and
Chakraborty, Tanmoy and
Solorio, Thamar and
Das, Amitava},
booktitle = {Proceedings of SemEval},
year = {2020},
}

@inproceedings{zampieri-etal-2019-semeval,
    title = "{S}em{E}val-2019 Task 6: Identifying and Categorizing Offensive Language in Social Media ({O}ffens{E}val)",
    author = "Zampieri, Marcos  and
      Malmasi, Shervin  and
      Nakov, Preslav  and
      Rosenthal, Sara  and
      Farra, Noura  and
      Kumar, Ritesh",
    booktitle = "Proceedings of SemEval",
    year = "2019",
}

@inproceedings{zampieri-etal-2020-semeval,
    title = "{S}em{E}val-2020 Task 12: Multilingual Offensive Language Identification in Social Media ({O}ffens{E}val 2020)",
    author = {Zampieri, Marcos  and
      Nakov, Preslav  and
      Rosenthal, Sara  and
      Atanasova, Pepa  and
      Karadzhov, Georgi  and
      Mubarak, Hamdy  and
      Derczynski, Leon  and
      Pitenis, Zeses  and
      {\c{C}}{\"o}ltekin, {\c{C}}a{\u{g}}r{\i}},
    booktitle = "Proceedings of SemEval",
    year = "2020",
}

@inproceedings{barbieri2016overview,
  title={Overview of the evalita 2016 sentiment polarity classification task},
  author={Barbieri, Francesco and Basile, Valerio and Croce, Danilo and Nissim, Malvina and Novielli, Nicole and Patti, Viviana and others},
  booktitle={CEUR Workshop Proceedings},
  year={2016},
}

@inproceedings{lee2024overview,
  title={Overview of the SIGHAN 2024 shared task for Chinese dimensional aspect-based sentiment analysis},
  author={Lee, Lung-Hao and Yu, Liang-Chih and Wang, Suge and Liao, Jian},
  booktitle={Proceedings of SIGHAN},
  year={2024}
}

@inproceedings{giorgi-etal-2024-findings,
    title = "Findings of {WASSA} 2024 Shared Task on Empathy and Personality Detection in Interactions",
    author = "Giorgi, Salvatore  and
      Sedoc, Jo{\~a}o  and
      Barriere, Valentin  and
      Tafreshi, Shabnam",
    booktitle = "Proceedings of WASSA",
    year = "2024",
}

@inproceedings{mohammad-bravo-marquez-2017-wassa,
    title = "{WASSA}-2017 Shared Task on Emotion Intensity",
    author = "Mohammad, Saif  and
      Bravo-Marquez, Felipe",
    booktitle = "Proceedings of WASSA",
    year = "2017",
}

@article{pan2024overview,
  title={Overview of EmoSPeech at IberLEF 2024: Multimodal Speech-text Emotion Recognition in Spanish},
  author={Pan, Ronghao and Garc{\'\i}a-D{\'\i}az, Jos{\'e} Antonio and Rondr{\'\i}guez-Garc{\'\i}a, Miguel {\'A}ngel and Garc{\'\i}a-S{\'a}nchez, Francisco and Valencia-Garc{\'\i}a, Rafael},
  journal={Procesamiento del Lenguaje Natural},
  volume={73},
  pages={359--368},
  year={2024}
}

@inproceedings{balouchzahi-etal-2022-overview,
    title = "Overview of {C}o{LI}-Kanglish: Word Level Language Identification in Code-mixed {K}annada-{E}nglish Texts at {ICON} 2022",
    author = "Balouchzahi, F.  and
      Butt, S.  and
      Hegde, A.  and
      Ashraf, N.  and
      Shashirekha, H.l.  and
      Sidorov, Grigori  and
      Gelbukh, Alexander",
    booktitle = "Proceedings of ICON",
    year = "2022",
}

@inproceedings{chakravarthi2023overview,
  title={Overview of the shared task on sarcasm identification of Dravidian languages (Malayalam and Tamil) in DravidianCodeMix},
  author={Chakravarthi, Bharathi Raja and Sripriya, N and Bharathi, B and Nandhini, K and Navaneethakrishnan, S Chinnaudayar and Durairaj, Thenmozhi and Ponnusamy, Rahul and Kumaresan, Prasanna Kumar and Ponnusamy, Kishore Kumar and Rajkumar, Charmathi},
  booktitle={Proceedings of FIRE},
  year={2023}
}

@inproceedings{chakravarthi2020overview,
  title={Overview of the track on sentiment analysis for dravidian languages in code-mixed text},
  author={Chakravarthi, Bharathi Raja and Priyadharshini, Ruba and Muralidaran, Vigneshwaran and Suryawanshi, Shardul and Jose, Navya and Sherly, Elizabeth and McCrae, John P},
  booktitle={Proceedings of FIRE},
  year={2020}
}

@inproceedings{satapara2021overview,
  title={Overview of the HASOC Subtrack at FIRE 2021: Conversational Hate Speech Detection in Code-mixed language.},
  author={Satapara, Shrey and Modha, Sandip and Mandl, Thomas and Madhu, Hiren and Majumder, Prasenjit},
  booktitle={FIRE (Working Notes)},
  year={2021}
}

@inproceedings{modha2022overview,
  title={Overview of the HASOC Subtrack at FIRE 2022: Identification of Conversational Hate-Speech in Hindi-English Code-Mixed and German Language.},
  author={Modha, Sandip and Mandl, Thomas and Majumder, Prasenjit and Satapara, Shrey and Patel, Tithi and Madhu, Hiren},
  booktitle={FIRE (Working Notes)},
  year={2022}
}

@inproceedings{calcs2018shtask,
    title={{Overview of the CALCS 2018 Shared Task: Named Entity Recognition on Code-switched Data}},
    author={Aguilar, Gustavo and AlGhamdi, Fahad and Soto, Victor and Diab, Mona and Hirschberg, Julia and Solorio, Thamar},
    booktitle = {Proceedings of CALCS},
    year={2018},
}

@inproceedings{sravani-etal-2021-political,
    title = "Political Discourse Analysis: A Case Study of Code Mixing and Code Switching in Political Speeches",
    author = "Sravani, Dama  and
      Kameswari, Lalitha  and
      Mamidi, Radhika",
    booktitle = "Proceedings of CALCS",
    year = "2021",
}

@inproceedings{yong-etal-2023-prompting,
    title = "Prompting Multilingual Large Language Models to Generate Code-Mixed Texts: The Case of South {E}ast {A}sian Languages",
    author = "Yong, Zheng Xin  and
      Zhang, Ruochen  and
      Forde, Jessica  and
      Wang, Skyler  and
      Subramonian, Arjun  and
      Lovenia, Holy  and
      Cahyawijaya, Samuel  and
      Winata, Genta  and
      Sutawika, Lintang  and
      Cruz, Jan Christian Blaise  and
      Tan, Yin Lin  and
      Phan, Long  and
      Phan, Long  and
      Garcia, Rowena  and
      Solorio, Thamar  and
      Aji, Alham Fikri",
    booktitle = "Proceedings of CALCS",
    year = "2023",
}

@inproceedings{malmasi-etal-2016-discriminating,
    title = "Discriminating between Similar Languages and {A}rabic Dialect Identification: A Report on the Third {DSL} Shared Task",
    author = {Malmasi, Shervin  and
      Zampieri, Marcos  and
      Ljube{\v{s}}i{\'c}, Nikola  and
      Nakov, Preslav  and
      Ali, Ahmed  and
      Tiedemann, J{\"o}rg},
    booktitle = "Proceedings of VarDial",
    year = "2016",
}

@inproceedings{zampieri-etal-2019-report,
    title = "A Report on the Third {V}ar{D}ial Evaluation Campaign",
    author = "Zampieri, Marcos  and
      Malmasi, Shervin  and
      Scherrer, Yves  and
      Samard{\v{z}}i{\'c}, Tanja  and
      Tyers, Francis  and
      Silfverberg, Miikka  and
      Klyueva, Natalia  and
      Pan, Tung-Le  and
      Huang, Chu-Ren  and
      Ionescu, Radu Tudor  and
      Butnaru, Andrei M.  and
      Jauhiainen, Tommi",
    booktitle = "Proceedings of VarDial",
    year = "2019",
}

@inproceedings{gaman-etal-2020-report,
    title = "A Report on the {V}ar{D}ial Evaluation Campaign 2020",
    author = "Gaman, Mihaela  and
      Hovy, Dirk  and
      Ionescu, Radu Tudor  and
      Jauhiainen, Heidi  and
      Jauhiainen, Tommi  and
      Lind{\'e}n, Krister  and
      Ljube{\v{s}}i{\'c}, Nikola  and
      Partanen, Niko  and
      Purschke, Christoph  and
      Scherrer, Yves  and
      Zampieri, Marcos",
    booktitle = "Proceedings of VarDial",
    year = "2020",
}

@inproceedings{chifu-etal-2024-vardial,
    title = "{V}ar{D}ial Evaluation Campaign 2024: Commonsense Reasoning in Dialects and Multi-Label Similar Language Identification",
    author = "Chifu, Adrian - Gabriel  and
      Glava{\v{s}}, Goran  and
      Ionescu, Radu Tudor  and
      Ljube{\v{s}}i{\'c}, Nikola  and
      Mileti{\'c}, Aleksandra  and
      Mileti{\'c}, Filip  and
      Scherrer, Yves  and
      Vuli{\'c}, Ivan",
    booktitle = "Proceedings of VarDial",
    year = "2024",
}

@inproceedings{abdul-mageed-etal-2024-nadi,
    title = "{NADI 2024: The Fifth Nuanced Arabic Dialect Identification Shared Task}",
    author = "Abdul - Mageed, Muhammad  and
      Keleg, Amr and
      Elmadany, AbdelRahim and
      Zhang, Chiyu  and
      Hamed, Injy and
      Magdy, Walid and
      Bouamor, Houda  and
      Habash, Nizar",
    booktitle = {Proceedings of ArabicNLP},
    year = "2024",
}

@inproceedings{abdul-mageed-etal-2023-nadi,
    title = "{NADI 2023: The Fourth Nuanced Arabic Dialect Identification Shared Task}",
    author = "Abdul - Mageed, Muhammad  and
      Elmadany, AbdelRahim and
      Zhang, Chiyu  and
      Nagoudi, ElMoatez Billah and
      Bouamor, Houda  and
      Habash, Nizar",
    booktitle = {Proceedings of ArabicNLP},
    year = "2023",
}

@inproceedings{abdul-mageed-etal-2022-nadi,
    title = "{NADI} 2022: The Third Nuanced {A}rabic Dialect Identification Shared Task",
    author = "Abdul - Mageed, Muhammad  and
      Zhang, Chiyu  and
      Elmadany, AbdelRahim  and
      Bouamor, Houda  and
      Habash, Nizar",
    booktitle = "Proceedings of WANLP",
    year = "2022",
}

@inproceedings{mageed:2021:nadi,
    title = "{NADI} 2021: The Second Nuanced Arabic Dialect Identification Shared Task",
    author = "Abdul-Mageed, Muhammad and Zhang, Chiyu and Elmadany, AbdelRahim and Bouamor, Houda and Habash, Nizar",
    booktitle = "Proceedings of WANLP",
    year = "2021",
}

@inproceedings{mageed-etal-2020-nadi,

    title ={{NADI 2020: The First Nuanced Arabic Dialect Identification Shared Task}},
    author = {Abdul - Mageed, Muhammad and Zhang, Chiyu and Bouamor, Houda and Habash, Nizar},
    booktitle ={Proceedings of WANLP},
    year = {2020},
}

@article{rabinovich2018native,
  title={Native language cognate effects on second language lexical choice},
  author={Rabinovich, Ella and Tsvetkov, Yulia and Wintner, Shuly},
  journal={Transactions of the Association for Computational Linguistics},
  volume={6},
  pages={329--342},
  year={2018},
  publisher={MIT Press One Rogers Street, Cambridge, MA 02142-1209, USA}
}

@inproceedings{ur2013cross,
  title={A cross-cultural framework for protecting user privacy in online social media},
  author={Ur, Blase and Wang, Yang},
  booktitle={Proceedings of WWW},
  year={2013}
}

@inproceedings{raghavan2012override,
  title={Override: A mobile privacy framework for context-driven perturbation and synthesis of sensor data streams},
  author={Raghavan, Kasturi Rangan and Chakraborty, Supriyo and Srivastava, Mani and Teague, Harris},
  booktitle={Proceedings of SenSys},
  year={2012}
}

@article{saka2024evaluating,
  title={Evaluating Privacy Measures in Healthcare Apps Predominantly Used by Older Adults},
  author={Saka, Suleiman and Das, Sanchari},
  journal={arXiv preprint arXiv:2410.14607},
  year={2024}
}

@inproceedings{bonneau2009privacy,
  title={Privacy suites: shared privacy for social networks.},
  author={Bonneau, Joseph and Anderson, Jonathan and Church, Luke},
  booktitle={Proceedings of SOUPS},
  year={2009}
}

@inproceedings{adu2008social,
  title={Social circles: Tackling privacy in social networks},
  author={Adu-Oppong, Fabeah and Gardiner, Casey K and Kapadia, Apu and Tsang, Patrick P},
  booktitle={Proceedings of SOUPS},
  year={2008}
}

@inproceedings{mondal2014understanding,
  title={Understanding and specifying social access control lists},
  author={Mondal, Mainack and Liu, Yabing and Viswanath, Bimal and Gummadi, Krishna P and Mislove, Alan},
  booktitle={Proceedings of SOUPS},
  year={2014}
}

@inproceedings{moore2024negative,
  title={Negative effects of social triggers on user security and privacy behaviors},
  author={Moore, Lachlan and Mori, Tatsuya and Hasegawa, Ayako A},
  booktitle={Proceedings of SOUPS},
  year={2024}
}

@article{voigt2017eu,
  title={The eu general data protection regulation (gdpr)},
  author={Voigt, Paul and Von dem Bussche, Axel},
  journal={A Practical Guide, 1st Ed., Cham: Springer International Publishing},
  volume={10},
  pages={10--5555},
  year={2017},
  publisher={Springer}
}

@article{goldman2020introduction,
  title={An introduction to the california consumer privacy act (ccpa)},
  author={Goldman, Eric},
  journal={Santa Clara Univ. Legal Studies Research Paper},
  year={2020}
}

@inproceedings{shokri2017membership,
  title={Membership inference attacks against machine learning models},
  author={Shokri, Reza and Stronati, Marco and Song, Congzheng and Shmatikov, Vitaly},
  booktitle={Proceedings of IEEE S\&P},
  year={2017},
  organization={IEEE}
}

@inproceedings{fredrikson2015model,
  title={Model inversion attacks that exploit confidence information and basic countermeasures},
  author={Fredrikson, Matt and Jha, Somesh and Ristenpart, Thomas},
  booktitle={Proceedings of CCS},
  year={2015}
}

@inproceedings{sharma2022named,
  title={Named entity recognition in natural language processing: A systematic review},
  author={Sharma, Abhishek and Amrita and Chakraborty, Sudeshna and Kumar, Shivam},
  booktitle={Proceedings of DoSCI},
  year={2022},
}

@inproceedings{chakravarthi2020sentiment,
  title={A Sentiment Analysis Dataset for Code-Mixed Malayalam-English},
  author={Chakravarthi, Bharathi Raja and Jose, Navya and Suryawanshi, Shardul and Sherly, Elizabeth and McCrae, John Philip},
  booktitle={Joint Proceedings of SLTU and CCURL},
  year={2020}
}

@article{mehta2022social,
  title={Social media hate speech detection using explainable artificial intelligence (XAI)},
  author={Mehta, Harshkumar and Passi, Kalpdrum},
  journal={Algorithms},
  volume={15},
  year={2022},
  publisher={MDPI}
}

@article{gangarde2021privacy,
  title={Privacy preservation in online social networks using multiple-graph-properties-based clustering to ensure k-anonymity, l-diversity, and t-closeness},
  author={Gangarde, Rupali and Sharma, Amit and Pawar, Ambika and Joshi, Rahul and Gonge, Sudhanshu},
  journal={Electronics},
  volume={10},
  year={2021},
  publisher={MDPI}
}

@article{alloghani2019systematic,
  title={A systematic review on the status and progress of homomorphic encryption technologies},
  author={Alloghani, Mohamed and Alani, Mohammed M and Al-Jumeily, Dhiya and Baker, Thar and Mustafina, Jamila and Hussain, Abir and Aljaaf, Ahmed J},
  journal={Journal of Information Security and Applications},
  volume={48},
  year={2019},
  publisher={Elsevier}
}

@article{mchugh2012interrater,
  title={Interrater reliability: the kappa statistic},
  author={McHugh, Mary L},
  journal={Biochemia medica},
  volume={22},
  year={2012},
  publisher={Medicinska naklada}
}

@article{mahendran2021privacy,
  title={Privacy-preservation in the context of natural language processing},
  author={Mahendran, Darshini and Luo, Changqing and Mcinnes, Bridget T},
  journal={IEEE Access},
  volume={9},
  pages={147600--147612},
  year={2021},
  publisher={IEEE}
}

@inproceedings{trac2-report,
  title={Evaluating aggression identification in social media},
  author={Kumar, Ritesh and Ojha, Atul and Malmasi, Shervin and Zampieri, Marcos},
  booktitle={Proceedings of TRAC},
  year={2020}
}

@inproceedings{lucas2008flybynight,
  title={Flybynight: mitigating the privacy risks of social networking},
  author={Lucas, Matthew M and Borisov, Nikita},
  booktitle={Proceedings of WPES},
  year={2008}
}

@inproceedings{zhang2018privacy,
  title={Privacy-preserving social media data outsourcing},
  author={Zhang, Jinxue and Sun, Jingchao and Zhang, Rui and Zhang, Yanchao and Hu, Xia},
  booktitle={Proceedings of INFOCOM},
  year={2018},
  organization={IEEE}
}

@article{bioglio2022analysis,
  title={Analysis and classification of privacy-sensitive content in social media posts},
  author={Bioglio, Livio and Pensa, Ruggero G},
  journal={EPJ Data Science},
  volume={11},
  pages={12},
  year={2022},
  publisher={Springer Berlin Heidelberg}
}

@article{khalil2024federated,
  title={Federated learning for privacy-preserving depression detection with multilingual language models in social media posts},
  author={Khalil, Samar Samir and Tawfik, Noha S and Spruit, Marco},
  journal={Patterns},
  volume={5},
  year={2024},
  publisher={Elsevier}
}

@article{yue2019survey,
  title={A survey of sentiment analysis in social media},
  author={Yue, Lin and Chen, Weitong and Li, Xue and Zuo, Wanli and Yin, Minghao},
  journal={Knowledge and information systems},
  volume={60},
  pages={617--663},
  year={2019},
  publisher={Springer}
}

@inproceedings{andalibi2020human,
  title={The human in emotion recognition on social media: Attitudes, outcomes, risks},
  author={Andalibi, Nazanin and Buss, Justin},
  booktitle={Proceedings of CHI},
  year={2020}
}

@article{ataei2022pars,
  title={Pars-OFF: a benchmark for offensive language detection on farsi social media},
  author={Ataei, Taha Shangipour and Darvishi, Kamyar and Javdan, Soroush and Pourdabiri, Amin and Minaei-Bidgoli, Behrouz and Pilehvar, Mohammad Taher},
  journal={IEEE Transactions on Affective Computing},
  volume={14},
  pages={2787--2795},
  year={2022},
  publisher={IEEE}
}

@article{das2013code,
  title={Code-mixing in social media text},
  author={Das, Amitava and Gamb{\"a}ck, Bj{\"o}rn},
  journal={Traitement Automatique des Langues},
  volume={54},
  pages={41--64},
  year={2013}
}

@article{li2020review,
  title={A review of applications in federated learning},
  author={Li, Li and Fan, Yuxi and Tse, Mike and Lin, Kuo-Yi},
  journal={Computers \& Industrial Engineering},
  volume={149},
  year={2020},
  publisher={Elsevier}
}

@article{xu2025privacy,
  title={Privacy-preserving multimodal sentiment analysis},
  author={Xu, Honghui and Li, Wei and Takabi, Daniel and Seo, Daehee and Cai, Zhipeng},
  journal={IEEE Internet of Things Journal},
  year={2025},
  publisher={IEEE}
}

@article{silva2022privacy,
  title={Privacy risk assessment and privacy-preserving data monitoring},
  author={Silva, Paulo and Gon{\c{c}}alves, Carolina and Antunes, Nuno and Curado, Marilia and Walek, Bogdan},
  journal={Expert Systems with Applications},
  volume={200},
  pages={116867},
  year={2022},
  publisher={Elsevier}
}

@article{cho2020privacy,
  title={Privacy risks, emotions, and social media: A coping model of online privacy},
  author={Cho, Hichang and Li, Pengxiang and Goh, Zhang Hao},
  journal={ACM Transactions on Computer-Human Interaction (TOCHI)},
  volume={27},
  pages={1--28},
  year={2020},
  publisher={ACM New York, NY, USA}
}

@inproceedings{pan2020privacy,
  title={Privacy risks of general-purpose language models},
  author={Pan, Xudong and Zhang, Mi and Ji, Shouling and Yang, Min},
  booktitle={Proceedings of IEEE S\&P},
  year={2020},
}

@article{roy2025ensuring,
  title={Ensuring safety in digital spaces: Detecting code-mixed hate speech in social media posts},
  author={Roy, Pradeep Kumar and Kumar, Abhinav},
  journal={Data \& Knowledge Engineering},
  volume={156},
  pages={102409},
  year={2025},
  publisher={Elsevier}
}

@inproceedings{dai-etal-2025-stealing,
    title = "Stealing Training Data from Large Language Models in Decentralized Training through Activation Inversion Attack",
    author = "Dai, Chenxi  and
      Lu, Lin  and
      Zhou, Pan",
    booktitle = "Proceedings of ACL",
    year = "2025",
}

@inproceedings{meng2025rr,
  title={Rr: Unveiling llm training privacy through recollection and ranking},
  author={Meng, Wenlong and Zhenyuan, Guo and Wu, Lenan and Gong, Chen and Liu, Wenyan and Li, Weixian and Wei, Chengkun and Chen, Wenzhi},
  booktitle={Findings of ACL},
  year={2025}
}

@inproceedings{sutton2025named,
  title={Named Entity Inference Attacks on Clinical LLMs: Exploring Privacy Risks and the Impact of Mitigation Strategies},
  author={Sutton, Adam and Bai, Xi and Noor, Kawsar and Searle, Thomas and Dobson, Richard},
  booktitle={Proceedings of PrivateNLP},
  year={2025}
}

@inproceedings{ye2025llms,
  title={LLMs are Privacy Erasable},
  author={Ye, Zipeng and Luo, Wenjian},
  booktitle={Findings of EMNLP},
  year={2025}
}

@inproceedings{mattern-etal-2023-membership,
    title = "Membership Inference Attacks against Language Models via Neighbourhood Comparison",
    author = {Mattern, Justus  and
      Mireshghallah, Fatemehsadat  and
      Jin, Zhijing  and
      Sch{\"o}lkopf, Bernhard  and
      Sachan, Mrinmaya  and
      Berg-Kirkpatrick, Taylor},
    booktitle = "Findings of ACL",
    year = "2023",
}

@inproceedings{song-etal-2025-multilingual,
    title = "Multilingual Blending: Large Language Model Safety Alignment Evaluation with Language Mixture",
    author = "Song, Jiayang  and
      Huang, Yuheng  and
      Zhou, Zhehua  and
      Ma, Lei",
    booktitle = "Findings of NAACL",
    year = "2025",
}

@inproceedings{kim-etal-2024-pruning,
    title = "Pruning Multilingual Large Language Models for Multilingual Inference",
    author = "Kim, Hwichan  and
      Suzuki, Jun  and
      Hirasawa, Tosho  and
      Komachi, Mamoru",
    booktitle = "Findings of EMNLP",
    year = "2024",
}

@article{adhikaripolicypulse,
  title={PolicyPulse: Precision Semantic Role Extraction for Enhanced Privacy Policy Comprehension},
  author={Adhikari, Andrick and Das, Sanchari and Dewri, Rinku},
  journal={Proceedings of NDSS},
  year={2025}

}

@inproceedings{adhikari2022privacy,
  title={Privacy policy analysis with sentence classification},
  author={Adhikari, Andrick and Das, Sanchari and Dewri, Rinku},
  booktitle={2022 19th Annual International Conference on Privacy, Security \& Trust (PST)},
  pages={1--10},
  year={2022},
  organization={IEEE}
}

@article{adhikari2023evolution,
  title={Evolution of Composition, Readability, and Structure of Privacy Policies over Two Decades},
  author={Adhikari, Andrick and Das, Sanchari and Dewri, Rinku},
  journal={Proceedings on Privacy Enhancing Technologies},
  volume={3},
  pages={138{\'s}153},
  year={2023}
}

@inproceedings{das2022sok,
  title={SoK: a proposal for incorporating accessible gamified cybersecurity awareness training informed by a systematic literature review},
  author={Das, Sanchari and others},
  booktitle={Proceedings of the workshop on usable security and privacy (USEC)},
  year={2022}
}

@inproceedings{das2020change,
  title={Change-Point Analysis of Cyberbullying-Related Twitter Discussions During COVID-19},
  author={Das, Sanchari and Kim, Andrew and Karmakar, Sayar},
  booktitle={Proceedings of the 16th Annual Social Informatics Research Symposium (“Sociotechnical Change Agents: ICTs, Sustainability, and Global Challenges”) in Conjunction with the 83rd Association for Information Science and Technology (ASIS\&T)},
  year={2020}
}

@article{das2021does,
  title={Does this photo make me look good? how posters, outsiders, and friends evaluate social media photo posts},
  author={Das, Sanchari and Ahmed, Tousif and Kapadia, Apu and Patil, Sameer},
  journal={Proceedings of the ACM on Human-Computer Interaction},
  volume={5},
  number={CSCW1},
  pages={1--32},
  year={2021},
  publisher={ACM New York, NY, USA}
}

@article{shrestha2022exploring,
  title={Exploring gender biases in ML and AI academic research through systematic literature review},
  author={Shrestha, Sunny and Das, Sanchari},
  journal={Frontiers in artificial intelligence},
  volume={5},
  pages={976838},
  year={2022},
  publisher={Frontiers Media SA}
}

@inproceedings{das2019all,
  title={All About Phishing Exploring User Research through a Systematic Literature Review},
  author={Das, Sanchari and Kim, Andrew and Tingle, Zachary and Nippert-Eng, Christena},
  booktitle={Proceedings of the Thirteenth International Symposium on Human Aspects of Information Security \& Assurance (HAISA 2019)},
  year={2019}
}

@article{noah2021exploring,
  title={Exploring evolution of augmented and virtual reality education space in 2020 through systematic literature review},
  author={Noah, Naheem and Das, Sanchari},
  journal={Computer Animation and Virtual Worlds},
  volume={32},
  number={3-4},
  pages={e2020},
  year={2021},
  publisher={Wiley Online Library}
}

@inproceedings{jones2021literature,
  title={A literature review on virtual reality authentication},
  author={Jones, John M and Duezguen, Reyhan and Mayer, Peter and Volkamer, Melanie and Das, Sanchari},
  booktitle={International Symposium on Human Aspects of Information Security and Assurance},
  pages={189--198},
  year={2021},
  organization={Springer}
}

@inproceedings{kishnani2023blockchain,
  title={Blockchain in oil and gas supply chain: a literature review from user security and privacy perspective},
  author={Kishnani, Urvashi and Madabhushi, Srinidhi and Das, Sanchari},
  booktitle={International Symposium on Human Aspects of Information Security and Assurance},
  pages={296--309},
  year={2023},
  organization={Springer}
}

@inproceedings{agarwal2025systematic,
  title={Systematic Literature Review of Vulnerabilities and Defenses in VPNs, Tor, and Web Browsers},
  author={Agarwal, Neha and Mackin, Ethan and Tazi, Faiza and Grover, Mayank and More, Rutuja and Das, Sanchari},
  booktitle={International Conference on Information Systems Security},
  pages={357--375},
  year={2025},
  organization={Springer}
}

@inproceedings{majumdar2021sok,
  title={Sok: An evaluation of quantum authentication through systematic literature review},
  author={Majumdar, Ritajit and Das, Sanchari},
  booktitle={Proceedings of the Workshop on Usable Security and Privacy (USEC)},
  year={2021}
}

@inproceedings{podapati2025sok,
  title={SoK: a systematic review of context-and behavior-aware adaptive authentication in mobile environments},
  author={Podapati, Vyoma Harshitha and Nigam, Divyansh and Das, Sanchari},
  booktitle={International Symposium on Human Aspects of Information Security and Assurance},
  pages={406--419},
  year={2025},
  organization={Springer}
}

@article{saka2025sok,
  title={SoK: Reviewing Two Decades of Security, Privacy, Accessibility, and Usability Studies on Internet of Things for Older Adults},
  author={Saka, Suleiman and Das, Sanchari},
  journal={arXiv preprint arXiv:2512.16394},
  year={2025}
}

@article{tazi2024sok,
  title={Sok: Analyzing privacy and security of healthcare data from the user perspective},
  author={Tazi, Faiza and Nandakumar, Archana and Dykstra, Josiah and Rajivan, Prashanth and Das, Sanchari},
  journal={ACM Transactions on Computing for Healthcare},
  volume={5},
  number={2},
  pages={1--31},
  year={2024},
  publisher={ACM New York, NY}
}

@inproceedings{grover2025sok,
  title={SoK: a systematic review of privacy and security in healthcare robotics},
  author={Grover, Mayank and Das, Sanchari},
  booktitle={International Conference on Social Robotics},
  pages={212--234},
  year={2025},
  organization={Springer}
}

@article{tazi2022sok,
  title={Sok: An evaluation of the secure end user experience on the dark net through systematic literature review},
  author={Tazi, Faiza and Shrestha, Sunny and De La Cruz, Junibel and Das, Sanchari},
  journal={Journal of Cybersecurity and Privacy},
  volume={2},
  number={2},
  pages={329--357},
  year={2022},
  publisher={MDPI}
}

@article{tazi2023sok,
  title={Sok: Analysis of user-centered studies focusing on healthcare privacy \& security},
  author={Tazi, Faiza and Nandakumar, Archana and Dykstra, Josiah and Rajivan, Prashanth and Das, Sanchari},
  journal={arXiv preprint arXiv:2306.06033},
  year={2023}
}

@inproceedings{zezulak2023sok,
  title={SoK: Evaluating Privacy and Security Concerns of Using Web Services for the Disabled Population},
  author={Zezulak, Alisa and Tazi, Faiza and Das, Sanchari},
  booktitle={7th Workshop on Technology and Consumer Protection (ConPro’23)},
  year={2023}
}

@inproceedings{huang2025systemization,
  title={Systemization of Knowledge (SoK): Goals, coverage, and evaluation in cybersecurity and privacy games},
  author={Huang, Yue and Grobler, Marthie and Ferro, Lauren S and Psaroulis, Georgia and Das, Sanchari and Wei, Jing and Janicke, Helge},
  booktitle={Proceedings of the 2025 CHI Conference on Human Factors in Computing Systems},
  pages={1--27},
  year={2025}
}

@inproceedings{duzgun2022sok,
  title={Sok: A systematic literature review of knowledge-based authentication on augmented reality head-mounted displays},
  author={D{\"u}zg{\"u}n, Reyhan and Noah, Naheem and Mayer, Peter and Das, Sanchari and Volkamer, Melanie},
  booktitle={Proceedings of the 17th International Conference on Availability, Reliability and Security},
  pages={1--12},
  year={2022}
}

@inproceedings{shrestha2022sok,
  title={SoK: A systematic literature review of bluetooth security threats and mitigation measures},
  author={Shrestha, Sunny and Irby, Esa and Thapa, Raghav and Das, Sanchari},
  booktitle={International Symposium on Emerging Information Security and Applications},
  pages={108--127},
  year={2022},
  organization={Springer}
}

@inproceedings{choi-etal-2025-safeguarding,
    title = "Safeguarding Privacy of Retrieval Data against Membership Inference Attacks: Is This Query Too Close to Home?",
    author = "Choi, Yujin  and
      Park, Youngjoo  and
      Byun, Junyoung  and
      Lee, Jaewook  and
      Park, Jinseong",
    booktitle = "Findings of EMNLP",
    year = "2025",
}

@article{das2025security,
  title={Security and privacy challenges of large language models: A survey},
  author={Das, Badhan Chandra and Amini, M Hadi and Wu, Yanzhao},
  journal={ACM Computing Surveys},
  volume={57},
  pages={1--39},
  year={2025},
  publisher={ACM New York, NY}
}

@inproceedings{acharya2025tracing,
  title={Tracing L1 Interference in English Learner Writing: A Longitudinal Corpus with Error Annotations},
  author={Acharya, Poorvi and Liebl, J Elizabeth and Goswami, Dhiman and North, Kai and Zampieri, Marcos and Anastasopoulos, Antonios},
  booktitle={Proceedings of EMNLP},
  year={2025}
}

@inproceedings{goswami2025multilingual,
  title={Multilingual Native Language Identification with Large Language Models},
  author={Goswami, Dhiman and Zampieri, Marcos and North, Kai and Malmasi, Shervin and Anastasopoulos, Antonios},
  booktitle={Proceedings of NAACL (SRW)},
  pages={193--199},
  year={2025}
}

@inproceedings{noman2019techies,
  title={Techies against Facebook: understanding negative sentiment toward Facebook via user generated content},
  author={Noman, Abu Saleh Md and Das, Sanchari and Patil, Sameer},
  booktitle={Proceedings of CHI},
  year={2019}
}

@inproceedings{danescu2011mark,
  title={Mark my words! Linguistic style accommodation in social media},
  author={Danescu-Niculescu-Mizil, Cristian and Gamon, Michael and Dumais, Susan},
  booktitle={Proceedings of WWW},
  pages={745--754},
  year={2011}
}

@article{yanushkevich2018understanding,
  title={Understanding and taxonomy of uncertainty in modeling, simulation, and risk profiling for border control automation},
  author={Yanushkevich, Svetlana N and Eastwood, Shawn C and Drahansky, Martin and Shmerko, Vlad P},
  journal={The Journal of Defense Modeling and Simulation},
  volume={15},
  pages={95--109},
  year={2018},
}

@article{liu2023cultural,
  title={Cultural bias in large language models: A comprehensive analysis and mitigation strategies},
  author={Liu, Zhaoming},
  journal={Journal of Transcultural Communication},
  volume={3},
  number={2},
  pages={224--244},
  year={2023},
  publisher={De Gruyter}
}

@inproceedings{zafarani2013connecting,
  title={Connecting users across social media sites: a behavioral-modeling approach},
  author={Zafarani, Reza and Liu, Huan},
  booktitle={Proceedings of SIGKDD},
  pages={41--49},
  year={2013}
}

@article{mai2016marginalization,
  title={Marginalization and Exclusion: Unraveling Systemic Bias in Classification.},
  author={Mai, Jens-Erik},
  journal={Knowledge Organization},
  volume={43},
  number={5},
  year={2016}
}

@inproceedings{mei2017inference,
  title={Inference attacks based on neural networks in social networks},
  author={Mei, Bo and Xiao, Yinhao and Li, Hong and Cheng, Xiuzhen and Sun, Yunchuan},
  booktitle={Proceedings of HotWeb},
  pages={1--6},
  year={2017}
}

@article{raza2025detecting,
  title={Detecting hate in diversity: a survey of multilingual code-mixed image and video analysis},
  author={Raza Ur Rehman, Hafiz Muhammad and Saleem, Mahpara and Jhandir, Muhammad Zeeshan and Alvarado, Eduardo Silva and Garay, Helena and Ashraf, Imran},
  journal={Journal of Big Data},
  volume={12},
  year={2025},
  publisher={Springer}
}

\section{Venue and Task-wise Paper Selection Details}

\begin{table}[!htp]
\centering

\resizebox{\linewidth}{!}{
\begin{tabular}{lcc}
\hline
\textbf{Venue} & \textbf{Initial Count} & \textbf{Final Count} \\
\hline
NDSS & 16 & 1 \\
CCS & 68 & 2\\
USENIX & 689 & 3 \\
S\&P & 19 & 3\\
SOUPS & 19 & 4\\
CSCW & 43 & 5\\
ICWSM & 71 & 7\\
CHI & 165 & 8\\
EMNLP & 1513 & 36\\
ACL & 1272 & 41\\
Other Journals and Conferences & 107 & 93\\
\hline
\textbf{Total} & \textbf{3982} & \textbf{203}\\
\hline
\end{tabular}
}

\caption{Venue-wise \# of the Publications Collected}
\label{tab:list1}
\end{table}

\begin{table}[!htp]
\centering
\resizebox{\linewidth}{!}{%
\begin{tabular}{lcc}
\hline
\textbf{Task} & \textbf{Paper Count} \\
\hline
Sentiment Analysis & 16 \\
Emotion Detection & 14 \\
Offensive Language Identification & 19 \\
Code-Mixing & 39 \\
Native Language Identification & 29 \\
Dialect Identification & 24 \\
Privacy & 62 \\

\hline
\textbf{Total} & \textbf{203}\\
\hline
\end{tabular}
}

\caption{Task-wise \# of the Peer-reviewed Publications}
\label{tab:list2}
\end{table}

\section{Additional Survey Findings and Experimental Results}
\label{appendix:DBC}

Table~\ref{tab:list4} shows the survey results of Datasets, Benchamark Competitions, and Computational Methodology in Social Media NLP Tasks.

Table \ref{tab:results1} shows Accuracy, Precision and Recall across all the models.

\begin{table*}[!htp]
\centering

\resizebox{\linewidth}{!}{%
\begin{tabular}{p{3.55cm}p{7.7cm}p{7.5cm}p{12cm}}
\hline
\textbf{Tasks} & \textbf{Datasets} & \textbf{Competitions} & \textbf{Computational Methodology}\\
\hline
\textbf{Sentiment Analysis} & 
SentMix-3L~\cite{raihan2023sentmix}, 

Bambara-French~\cite{konate2018sentiment}, 

Yelp~\cite{zhang2015character}, 

Sentiment140~\cite{go2009twitter}  & SemEval~\cite{muhammad-etal-2023-semeval,barnes-etal-2022-semeval,patwa2020sentimix},

EvalITA~\cite{barbieri2016overview},

SIGHAN~\cite{lee2024overview} & 
Feature Engineering, Deep Learning, Word2Vec~\cite{barbieri2016overview}, 

Dependency Graph~\cite{muhammad-etal-2023-semeval}, 

Transformer~\cite{lee2024overview,muhammad-etal-2023-semeval,barnes-etal-2022-semeval,patwa2020sentimix,raihan2023sentmix}, 

LLM Prompting~\cite{lee2024overview, raihan2023sentmix}, 

LSTM~\cite{konate2018sentiment}, 

CNN, SVM, Naive Bayes~\cite{barbieri2016overview,konate2018sentiment}, 

LAPT, TAPT, Ensembles, Sentence Transformer~\cite{barnes-etal-2022-semeval}\\
\hline
\textbf{Emotion Detection} & 
EmoMix-3L~\cite{raihan2024emomix}, 

MELD~\cite{poria2019meld}, 

ArtELingo~\cite{mohamed2022artelingo}, 

EmoReact~\cite{nojavanasghari2016emoreact} & SemEval~\cite{wang2024SemEval,kumar-etal-2024-semeval},

WASSA~\cite{giorgi-etal-2024-findings,mohammad-bravo-marquez-2017-wassa}, 

IberLEF~\cite{pan2024overview} & 
SVM, LSTM~\cite{wang2024SemEval,mohammad-bravo-marquez-2017-wassa}, 

LLM Fine-tuning~\cite{kumar-etal-2024-semeval}, 

Transformer~\cite{kumar-etal-2024-semeval,giorgi-etal-2024-findings,pan2024overview,raihan2024emomix}, 

LLM Prompting~\cite{wang2024SemEval,giorgi-etal-2024-findings,raihan2024emomix}, 

CNN, Word2Vec~\cite{mohammad-bravo-marquez-2017-wassa}, 

LoRA~\cite{pan2024overview}\\
\hline
\textbf{Offensive Language Identification} & 

OffMix-3L~\cite{goswami2023offmix}, 

COLD~\cite{deng-etal-2022-cold}, 

RUHSOLD~\cite{dewani2023detection}, 

KOLD~\cite{jeong-etal-2022-kold}, 

$DK_{HATE}$~\cite{sigurbergsson-derczynski-2020-offensive}, 

OGTD~\cite{pitenis-etal-2020-offensive}, 

OLID~\cite{zampieri2019predicting}, 

SOLID~\cite{rosenthal2021solid}, 

OLID-BR~\cite{trajano2024olid}, 

TB-OLID~\cite{raihan2023offensive} & SemEval~\cite{zampieri-etal-2019-semeval,zampieri-etal-2020-semeval}, 

HASOC~\cite{ghosh2023annihilate,ranasinghe2022overview,mandl2019overview,mandl2020overview}, 

TRAC~\cite{kumar-etal-2018-benchmarking,trac2-report},

GermEval~\cite{risch-etal-2021-overview}, 

DravidianLangTech~\cite{ravikiran-etal-2022-findings, chakravarthi-etal-2021-findings-shared, toxicspans-acl,premjith-2024-eacl-hate-telugu} & 

Bi-LSTM~\cite{zampieri-etal-2019-semeval,ghosh2023annihilate,premjith-2024-eacl-hate-telugu}, 

Transformer~\cite{goswami2023offmix,premjith-2024-eacl-hate-telugu,zampieri-etal-2019-semeval,zampieri-etal-2020-semeval,mandl2019overview,ranasinghe2022overview,trac2-report,ravikiran-etal-2022-findings, chakravarthi-etal-2021-findings-shared},

LLM Prompting~\cite{goswami2023offmix}, 

LR~\cite{ghosh2023annihilate,premjith-2024-eacl-hate-telugu,chakravarthi-etal-2021-findings-shared}, 

Naive Bayes~\cite{ghosh2023annihilate}, 

SVM~\cite{zampieri-etal-2019-semeval,ghosh2023annihilate,mandl2020overview,kumar-etal-2018-benchmarking,trac2-report,premjith-2024-eacl-hate-telugu}, 

n-gram~\cite{mandl2020overview}, 

KNN, Deep Learning~\cite{ranasinghe2022overview}, 

CNN~\cite{kumar-etal-2018-benchmarking,trac2-report,zampieri-etal-2019-semeval,ghosh2023annihilate,premjith-2024-eacl-hate-telugu}, 

LSTM, TF-IDF~\cite{chakravarthi-etal-2021-findings-shared,kumar-etal-2018-benchmarking,trac2-report,mandl2020overview}, 

RNN~\cite{premjith-2024-eacl-hate-telugu}\\
\hline
\textbf{Code-Mixing} & 

English-Manipuri~\cite{lamabam2016language}, 

Indonesian-English~\cite{nabila2022analysis}, 

English-Kannada~\cite{lakshmi2017automatic}, 

English-Cameroon-French~\cite{tabe2023code}, 

Sinhala-English~\cite{shanmugalingam2019language}, 

AR-EN CS LID Corpus~\cite{sabty2021language}, 

Hinglish, Tenglish, Tamlish~\cite{chandu2018code} & ICON~\cite{balouchzahi-etal-2022-overview},

FIRE~\cite{chakravarthi2020overview,chakravarthi2023overview}, 

CALCS~\cite{calcs2018shtask,sravani-etal-2021-political,yong-etal-2023-prompting}, 

HASOC~\cite{modha2022overview,satapara2021overview} & Transformer~\cite{balouchzahi-etal-2022-overview,chakravarthi2020overview,chakravarthi2023overview,modha2022overview,satapara2021overview}, 

TF-IDF~\cite{balouchzahi-etal-2022-overview,satapara2021overview,lakshmi2017automatic}, 

CNN~\cite{chakravarthi2020overview,satapara2021overview}, 

SVM~\cite{chakravarthi2023overview,calcs2018shtask,shanmugalingam2019language}, 

KNN, Bi-LSTM~\cite{chakravarthi2023overview}, 

LSTM~\cite{calcs2018shtask,balouchzahi-etal-2022-overview}, 

LLM Prompting~\cite{yong-etal-2023-prompting}, 

n-gram~\cite{lamabam2016language,lakshmi2017automatic}, 

BoW~\cite{lakshmi2017automatic}, 

LR, Decision Tree, Naive Bayes~\cite{shanmugalingam2019language}\\
\hline
\textbf{Native Language Identification} & Reddit-L2~\cite{rabinovich2018native}, 

INLI 2013~\cite{anand2017overview}, 

INLI 2017~\cite{soman2018overview} & NLI~\cite{tetreault2013report,malmasi2017report}, 

INLI~\cite{anand2017overview,soman2018overview} & 

n-gram~\cite{koppel2005determining}, 

TF-IDF~\cite{wong2009contrastive}, 

Lexical and Syntactic Feature~\cite{mechti2016empirical,gebre2013improving}, 

SVM~\cite{jarvis2013maximizing}, 

LR~\cite{di2020good}, 

Ensemble and Meta Classifier~\cite{zampieri2017native},

Transformer~\cite{steinbakken2020native}, 

LLM~\cite{zhang2023native}\\
\hline
\textbf{Dialect Identification} & AfriDial~\cite{sassi2024afridial}, 

MPCA~\cite{malmasi2015arabic}, 

North African Dialect~\cite{berrimi2020arabic}, 

TIMIT~\cite{zissman1996automatic}, 

TwitterAAE~\cite{blodgett2016demographic}, 

DIALECTBENCH~\cite{faisal-etal-2024-dialectbench}, 

ArchiMob~\cite{samardzic2016archimob}, 

ITDI-FDI~\cite{2022-findings-vardial} & VarDial~\cite{malmasi-etal-2016-discriminating,zampieri-etal-2019-report,gaman-etal-2020-report,chifu-etal-2024-vardial}, 

NADI~\cite{abdul-mageed-etal-2024-nadi,abdul-mageed-etal-2023-nadi,abdul-mageed-etal-2022-nadi,mageed:2021:nadi,mageed-etal-2020-nadi} & 

n-gram, TF-IDF~\cite{abdul-mageed-etal-2023-nadi,mageed-etal-2020-nadi,mageed:2021:nadi,abdul-mageed-etal-2022-nadi,gaman-etal-2020-report}, 

Ensemble~\cite{abdul-mageed-etal-2023-nadi,mageed-etal-2020-nadi,mageed:2021:nadi,abdul-mageed-etal-2022-nadi,malmasi-etal-2016-discriminating},

LLM Prompting~\cite{abdul-mageed-etal-2023-nadi,abdul-mageed-etal-2022-nadi,chifu-etal-2024-vardial}, 

Transformer~\cite{abdul-mageed-etal-2024-nadi,gaman-etal-2020-report,chifu-etal-2024-vardial,zampieri-etal-2019-report}, 

CNN, SVM~\cite{gaman-etal-2020-report,zampieri-etal-2019-report,malmasi-etal-2016-discriminating}, 

LoRA~\cite{chifu-etal-2024-vardial}, 

Naive Bayes~\cite{malmasi-etal-2016-discriminating,zampieri-etal-2019-report}, 

LSTM, Meta Classifier, Bi-LSTM~\cite{zampieri-etal-2019-report}
\\
\hline
\end{tabular}%
}

\caption{Survey Result of Datasets, Benchamark Competitions, and Computational Methodology in Social Media NLP Tasks}
\label{tab:list4}
\end{table*}

\begin{table*}[htp]
\centering

\resizebox{\textwidth}{!}{%
\begin{tabular}{l|ccc|ccc|ccc|ccc|ccc|ccc}
\hline
 & \multicolumn{9}{c|}{\textbf{Finetuning}} & \multicolumn{9}{c}{\textbf{Privacy-Preserving Finetuning}} \\
\cline{2-19}
\textbf{Tasks} & \multicolumn{3}{c|}{\textbf{XLM-R}} & \multicolumn{3}{c|}{\textbf{GPT-2}} & \multicolumn{3}{c|}{\textbf{FLAN-T5}} & \multicolumn{3}{c|}{\textbf{XLM-R}} & \multicolumn{3}{c|}{\textbf{GPT-2}} & \multicolumn{3}{c}{\textbf{FLAN-T5}} \\
\cline{2-19}
 & \textbf{Acc} & \textbf{Prec} & \textbf{Rec} & \textbf{Acc} & \textbf{Prec} & \textbf{Rec} & \textbf{Acc} & \textbf{Prec} & \textbf{Rec} & \textbf{Acc} & \textbf{Prec} & \textbf{Rec} & \textbf{Acc} & \textbf{Prec} & \textbf{Rec} & \textbf{Acc} & \textbf{Prec} & \textbf{Rec} \\
\hline
Sentiment Analysis & 0.84 & 0.84 & 0.84 & 0.86 & 0.86 & 0.86 & 0.50 & 0.25 & 0.50 & 0.84 & 0.84 & 0.84 & 0.85 & 0.85 & 0.85 & 0.50 & 0.25 & 0.50 \\
Emotion Detection & 0.64 & 0.60 & 0.64 & 0.63 & 0.58 & 0.63 & 0.59 & 0.58 & 0.59 & 0.61 & 0.57 & 0.61 & 0.60 & 0.56 & 0.60 & 0.48 & 0.23 & 0.48 \\
Offensive Language Identification & 0.84 & 0.84 & 0.84 & 0.86 & 0.86 & 0.86 & 0.79 & 0.79 & 0.79 & 0.82 & 0.82 & 0.82 & 0.84 & 0.84 & 0.84 & 0.72 & 0.72 & 0.72 \\
Code-Mixing & 0.65 & 0.62 & 0.65 & 0.62 & 0.58 & 0.62 & 0.64 & 0.67 & 0.64 & 0.61 & 0.57 & 0.61 & 0.58 & 0.54 & 0.58 & 0.63 & 0.65 & 0.63 \\
Native Language Identification & 0.58 & 0.58 & 0.58 & 0.50 & 0.55 & 0.50 & 0.20 & 0.41 & 0.20 & 0.38 & 0.44 & 0.38 & 0.32 & 0.40 & 0.32 & 0.17 & 0.13 & 0.17 \\
Dialect Identification & 0.67 & 0.62 & 0.67 & 0.60 & 0.56 & 0.60 & 0.50 & 0.36 & 0.50 & 0.64 & 0.60 & 0.64 & 0.50 & 0.49 & 0.50 & 0.47 & 0.31 & 0.47 \\
\hline

\end{tabular}
}

\caption{Experimental Results for Privacy-Preserving NLP (Accuracy, Precision, Recall)}
\label{tab:results1}
\end{table*}

\section{Social Media Privacy Risk Analysis}
\label{appendix:risk}
Tables~\ref{tab:result1},~\ref{tab:result2},~\ref{tab:result3},~\ref{tab:result4},~\ref{tab:result5}, and~\ref{tab:result6} show the detailed Analysis of Social Media Privacy Risks across different tasks.

\begin{table*}[htbp]
\centering

\small
\resizebox{\linewidth}{!}{%
\begin{tabular}{p{4.7cm} *{15}{p{0.45cm}}}
\hline
\multicolumn{1}{c}{} & \multicolumn{15}{c}{\textbf{Data Collection and Usage}} \\
\hline
\textbf{Reference} &  
\rotatebox{54}{User Opinion} & 
\rotatebox{54}{Demographic} &  
\rotatebox{54}{Emotion State} & 
\rotatebox{54}{Vulnerability Leakage} & 
\rotatebox{54}{Behavioral Pattern} & 
\rotatebox{54}{Toxic Content} & 
\rotatebox{54}{Harassment Risk} & 
\rotatebox{54}{Context Misinterpretation} & 
\rotatebox{54}{Mixed Context} &  
\rotatebox{54}{Identity Risk} & 
\rotatebox{54}{L1 Marking} & 
\rotatebox{54}{Region Marking} & 
\rotatebox{54}{Speech Sensitivity} & 
\rotatebox{54}{Ethnic Trace} &\\
\hline
~\citet{xiao2018mobile} & $\bullet$   & $\bullet$  & $\bullet$  & $\bullet$ & $\bullet$  & $\bullet$   & $\bullet$  & $\bullet$  & $\bullet$  & $\bullet$   &  & $\bullet$  &  $\bullet$ & \\
~\citet{sharma2020role} \\
~\citet{lohar2021irish} \\
\hline
~\citet{raihan2024emomix} & $\bullet$   & $\bullet$ & $\bullet$  & $\bullet$  & $\bullet$ & $\bullet$ & $\bullet$  & $\bullet$  & $\bullet$  & $\bullet$  & $\bullet$ & $\bullet$ & $\bullet$ & $\bullet$   \\
~\citet{teodorescu2023language}\\
~\citet{arango2024multifold}\\
~\citet{zampieri-etal-2024-federated}\\
\hline
~\citet{sigurbergsson-derczynski-2020-offensive} & $\bullet$    & $\bullet$ &  & $\bullet$ & $\bullet$ & $\bullet$ & $\bullet$ & $\bullet$ & $\bullet$  & $\bullet$ & $\bullet$ & $\bullet$ & $\bullet$ & $\bullet$  \\
~\citet{rosenthal2021solid}\\
~\citet{deng-etal-2022-cold}\\
\hline
~\citet{hidayatullah2022systematic} & $\bullet$ &  &  &  &  &  &  & $\bullet$ & $\bullet$ & $\bullet$ & $\bullet$ & $\bullet$ &   $\bullet$ & \\
\hline
~\citet{staicu2023bilingual} & $\bullet$   & $\bullet$ & $\bullet$ &  &  &  &  &$\bullet$  &  $\bullet$ & $\bullet$ & $\bullet$ & $\bullet$ & $\bullet$ &  \\
~\citet{bierner2001alternative}\\
~\citet{goldin2018native}\\
~\citet{jauhiainen2019automatic}\\
\hline
~\citet{jorgensen2015challenges} &  $\bullet$  & $\bullet$ & $\bullet$ & $\bullet$ &  & $\bullet$ &  & $\bullet$ & $\bullet$ & $\bullet$ & $\bullet$ & $\bullet$ & $\bullet$ & $\bullet$  \\
~\citet{huang2015improved}\\
~\citet{fleisig-etal-2024-linguistic}\\
~\citet{barot2024tonecheck}\\
\hline
\end{tabular}
}

\caption{Analysis of Social Media Privacy Concerns (Data Collection and Usage). Each row of reference is associated to this concern of the corresponding tasks mentioned in Table~\ref{tab:list3}.}
\label{tab:result1}
\end{table*}

\begin{table*}[htbp]
\centering

\small
\resizebox{\linewidth}{!}{%
\begin{tabular}{p{4.7cm} *{17}{p{0.4cm}}}
\hline
\multicolumn{1}{c}{} & \multicolumn{17}{c}{\textbf{Data Preprocessing and Anonymization}} \\
\hline
\textbf{Reference} &  
\rotatebox{54}{Name-Entity Leak} & 
\rotatebox{54}{Quote Leakage} & 
\rotatebox{54}{Stylistic Markers} & 
\rotatebox{54}{Emotion Cues} & 
\rotatebox{54}{Intensity Tag} & 
\rotatebox{54}{Mood Triggers} & 
\rotatebox{54}{Slang Trace} & 
\rotatebox{54}{Sensitive Pairing} & 
\rotatebox{54}{Script Leakage} & 
\rotatebox{54}{Phonetic Spelling} & 
\rotatebox{54}{Spelling Errors} & 
\rotatebox{54}{Grammatical Patterns} & 
\rotatebox{54}{Syntax Clues} & 
\rotatebox{54}{Accent Trace} & 
\rotatebox{54}{Word Uniqueness} & 
\rotatebox{54}{Phrasing Style} &\\
\hline
~\citet{lohar2021irish} &  & $\bullet$ & $\bullet$ & $\bullet$ & $\bullet$ & $\bullet$ & $\bullet$ & $\bullet$ &  & $\bullet$ & $\bullet$ & $\bullet$ & $\bullet$ &  & $\bullet$ & $\bullet$  \\
~\citet{xiao2018mobile}\\
\hline
~\citet{teodorescu2023language} & $\bullet$ & $\bullet$ & $\bullet$ & $\bullet$ & $\bullet$ & $\bullet$ & $\bullet$ & $\bullet$ & $\bullet$ &  & $\bullet$ & $\bullet$ & $\bullet$ & $\bullet$ & $\bullet$ & $\bullet$ \\
\hline
~\citet{leonardelli2021agreeing} & $\bullet$ & $\bullet$ & $\bullet$ & $\bullet$ & $\bullet$ &  & $\bullet$ & $\bullet$ &  &  & $\bullet$ & $\bullet$ &  &  & $\bullet$ & $\bullet$ \\
~\citet{jeong-etal-2022-kold}\\
~\citet{sigurbergsson-derczynski-2020-offensive}\\
~\citet{rosenthal2021solid}\\
\hline
~\citet{hidayatullah2022systematic} &  &  & $\bullet$ &  &  &  & $\bullet$ & $\bullet$ &  & $\bullet$ & $\bullet$ & $\bullet$ & $\bullet$ &  & $\bullet$ & $\bullet$ \\
\hline
~\citet{berzak-etal-2017-predicting} & $\bullet$ &  & $\bullet$ &  &  &  &  & $\bullet$ &  &  &  & $\bullet$ & $\bullet$ & $\bullet$ & & $\bullet$ \\
\hline
~\citet{malmasi2015arabic} & $\bullet$ & $\bullet$ & $\bullet$ & $\bullet$ & $\bullet$ & $\bullet$ & $\bullet$ & $\bullet$ &  & $\bullet$ & $\bullet$ & $\bullet$ & $\bullet$  & $\bullet$ & $\bullet$ & $\bullet$ \\
~\citet{ferragne2007automatic}\\
~\citet{mahmud2023cyberbullying}\\
\hline

\end{tabular}
}

\caption{Analysis of Social Media Privacy Concerns (Data Preprocessing and Anonymization). Each row of reference is associated to this concern of the corresponding tasks mentioned in Table~\ref{tab:list3}.}
\label{tab:result2}
\end{table*}

\begin{table*}[htbp]
\centering

\small
\resizebox{\linewidth}{!}{%
\begin{tabular}{p{3.85cm} *{17}{p{0.4cm}}}
\hline
\multicolumn{1}{c}{} & \multicolumn{17}{c}{\textbf{Data Visibility and User Profiling Risks}} \\
\hline
\textbf{Reference} &  
\rotatebox{54}{Sentiment Pattern} & 
\rotatebox{54}{Group Clustering} & 
\rotatebox{54}{Profile Inference} & 
\rotatebox{54}{Behavioral Modeling} & 
\rotatebox{54}{Profile Tracking} & 
\rotatebox{54}{Exposure Risk} & 
\rotatebox{54}{Group Tagging} & 
\rotatebox{54}{Group Inference} & 
\rotatebox{54}{Speech Style} & 
\rotatebox{54}{Cultural Assumption} & 
\rotatebox{54}{Educational Bias} & 
\rotatebox{54}{Forced Clustering} & 
\rotatebox{54}{Region Profiling} & 
\rotatebox{54}{Cultural Tagging} & 
\rotatebox{54}{Group Labelling} &
\rotatebox{54}{L1 Trace} & \\
\hline
~\citet{lohar2021irish} & $\bullet$ & $\bullet$ & $\bullet$ & $\bullet$ & $\bullet$ & $\bullet$ & $\bullet$ & $\bullet$ & $\bullet$ &  &  &  & $\bullet$ &  & $\bullet$ &   \\
~\citet{sharma2020role}\\
\hline
~\citet{leonardelli2021agreeing} & $\bullet$ & $\bullet$ & $\bullet$ & $\bullet$ &  & $\bullet$ & $\bullet$ & $\bullet$ & $\bullet$ & $\bullet$ &  & $\bullet$ & $\bullet$ & $\bullet$ & $\bullet$ &  \\
~\citet{rosenthal2021solid}\\
\hline
~\citet{hidayatullah2022systematic} & $\bullet$ &  &  &  &  &  &  &  & $\bullet$ & $\bullet$ &  &  &  &  & & $\bullet$ \\
~\citet{zhang2023multilingual}\\
\hline
~\citet{jauhiainen2019automatic} &  & $\bullet$ & $\bullet$ &  & $\bullet$ &  &  &  &  & $\bullet$ & $\bullet$ &  & $\bullet$ & $\bullet$ & & $\bullet$ \\
~\citet{goldin2018native}\\
~\citet{aoyama2024modeling}\\
\hline

\end{tabular}
}

\caption{Analysis of Social Media Privacy Concerns (Data Preprocessing and Anonymization). Each row of reference is associated to this concern of the corresponding tasks mentioned in Table~\ref{tab:list3}.}
\label{tab:result3}
\end{table*}

\begin{table*}[htbp]
\centering

\small
\resizebox{\linewidth}{!}{%
\begin{tabular}{p{3cm} *{12}{p{0.75cm}}}
\hline
\multicolumn{1}{c}{} & \multicolumn{12}{c}{\textbf{Computational Risks in NLP}} \\
\hline
\textbf{Reference} &  
\rotatebox{54}{Gradient Leakage} & 
\rotatebox{54}{Model Inversion} & 
\rotatebox{54}{Adversarial Attack} & 
\rotatebox{54}{Memory Retention} & 
\rotatebox{54}{Model Overfitting} & 
\rotatebox{54}{LLM Data Leakage} & 
\rotatebox{54}{Lossy Tokenization} & 
\rotatebox{54}{Embedding Noise} & 
\rotatebox{54}{Misclassification} & 
\rotatebox{54}{L1 Hallucination} & 
\rotatebox{54}{Dialect-Mixing} &\\
\hline

~\citet{arango2024multifold} &   &  & $\bullet$ &  & $\bullet$ & $\bullet$ & $\bullet$ & $\bullet$ &$\bullet$ &$\bullet$ & $\bullet$ \\
~\citet{xiao-etal-2024-toxicloakcn}\\
~\citet{morabito2024stop}\\
\hline
~\citet{mahmud2023cyberbullying} &   &  &  &  & $\bullet$  &  & $\bullet$ & $\bullet$ &$\bullet$  & & $\bullet$  \\
\hline

\end{tabular}
}

\caption{Analysis of Social Media Privacy Concerns (Computational Risks in NLP). Each row of reference is associated to this concern of the corresponding tasks mentioned in Table~\ref{tab:list3}.}
\label{tab:result4}
\end{table*}

\begin{table*}[htbp]
\centering

\small
\resizebox{\linewidth}{!}{%
\begin{tabular}{p{3.2cm} *{18}{p{0.4cm}}}
\hline
\multicolumn{1}{c}{} & \multicolumn{18}{c}{\textbf{Bias, Fairness, and Discrimination Risks}} \\
\hline
\textbf{Reference} &  
\rotatebox{54}{Polarity Skewness} & 
\rotatebox{54}{Cultural Bias} & 
\rotatebox{54}{Language Bias} & 
\rotatebox{54}{Labeling Bias} & 
\rotatebox{54}{Gender Bias} & 
\rotatebox{54}{Underrepresented Group} & 
\rotatebox{54}{Minority Flag} & 
\rotatebox{54}{Informality Bias} & 
\rotatebox{54}{Code-switching Flag} & 
\rotatebox{54}{Monolingual Bias} & 
\rotatebox{54}{Label Noise} & 
\rotatebox{54}{Context Loss} & 
\rotatebox{54}{Majority L2 Focus} & 
\rotatebox{54}{Underrepresented L1 Bias} & 
\rotatebox{54}{Sociolect Bias} & 
\rotatebox{54}{Wrong Annotation} & 
\rotatebox{54}{Accent Bias} &\\

\hline
~\citet{konate2018sentiment} & $\bullet$ &  & $\bullet$ & $\bullet$ &  & $\bullet$ &  & $\bullet$ & $\bullet$ & $\bullet$ & $\bullet$ & $\bullet$ & $\bullet$ & $\bullet$ & & $\bullet$ &   \\
~\citet{barbieri2016overview}\\
\hline
~\citet{fleisig-etal-2024-linguistic} &  & $\bullet$ & $\bullet$ & $\bullet$ & $\bullet$ & $\bullet$ & $\bullet$ & $\bullet$ &  & $\bullet$ & & $\bullet$ & $\bullet$ & $\bullet$  & $\bullet$ & & $\bullet$  \\
\hline

\end{tabular}
}

\caption{Analysis of Social Media Privacy Concerns (Bias, Fairness, and Discrimination Risks). Each row of reference is associated to this concern of the corresponding tasks mentioned in Table~\ref{tab:list3}.}
\label{tab:result5}
\end{table*}

\begin{table*}[htbp]
\centering

\small
\resizebox{\linewidth}{!}{%
\begin{tabular}{p{3.25cm} *{16}{p{0.45cm}}}
\hline
\multicolumn{1}{c}{} & \multicolumn{16}{c}{\textbf{Regulatory Compliance and Ethics}} \\
\hline
\textbf{Reference} &
\rotatebox{54}{Data Reuse} & 
\rotatebox{54}{Purpose Mismatch} & 
\rotatebox{54}{Right to be Forgotten} & 
\rotatebox{54}{HIPAA} & 
\rotatebox{54}{Longitudinal Use} & 
\rotatebox{54}{Defamation Liability} & 
\rotatebox{54}{Ethical Data Review} & 
\rotatebox{54}{Data Consent} & 
\rotatebox{54}{Speech Control} & 
\rotatebox{54}{Transparency} & 
\rotatebox{54}{Local Rights} & 
\rotatebox{54}{Opt in-out policy} & 
\rotatebox{54}{Multilingual Consent} &
\rotatebox{54}{GDPR} & 
\rotatebox{54}{CCPA} &\\

\hline
~\citet{xiao2018mobile} &$\bullet$  & $\bullet$ &  &  &  &  &$\bullet$  &$\bullet$  &  & $\bullet$ &  & $\bullet$ & $\bullet$ &  &   \\
~\citet{raihan2023sentmix}\\
\hline
~\citet{mohamed2022artelingo} & $\bullet$ &  &  & $\bullet$ & $\bullet$ &  & $\bullet$ & $\bullet$ & $\bullet$ & $\bullet$ & $\bullet$ & $\bullet$ & $\bullet$ & $\bullet$ & $\bullet$  \\
~\citet{raihan2024emomix}\\
~\citet{teodorescu2023language}\\
\hline
~\citet{zampieri-etal-2024-federated} &$\bullet$  &  & $\bullet$ &  &  &$\bullet$  &$\bullet$  &$\bullet$  &$\bullet$  & $\bullet$ & $\bullet$ & $\bullet$ &  & $\bullet$ &   \\
~\citet{jeong-etal-2022-kold}\\
\hline
~\citet{hidayatullah2022systematic} &$\bullet$  & $\bullet$ &  &  &  &  &$\bullet$  &  & $\bullet$ &  &  &  &  &  &   \\
~\citet{garg2018code} \\
\hline
~\citet{jauhiainen2019automatic} & $\bullet$ & $\bullet$ &  &  &  &  & $\bullet$ &  & $\bullet$ & $\bullet$ &  &  &  &  &   \\
~\citet{nguyen2021automatic}\\
\hline
~\citet{huang2015improved} &$\bullet$  &  &  &  &  &  & $\bullet$ & $\bullet$ &  & $\bullet$ & $\bullet$ & $\bullet$ &$\bullet$  & $\bullet$ & $\bullet$ \\
~\citet{fleisig-etal-2024-linguistic}\\
~\citet{faisal-etal-2024-dialectbench}\\
~\citet{barot2024tonecheck}\\
\hline

\end{tabular}
}

\caption{Analysis of Social Media Privacy Concerns (Regulatory Compliance and Ethics). Each row of reference is associated to this concern of the corresponding tasks mentioned in Table~\ref{tab:list3}.}
\label{tab:result6}
\end{table*}

\section{Social Media Privacy Risk Mitigation Analysis}
Table~\ref{tab:result7} shows the detailed analysis of mitigation strategies of privacy concerns in social media.
\begin{table*}[htbp]
\centering

\small
\resizebox{\linewidth}{!}{%
\begin{tabular}{p{5.5cm} *{15}{p{0.35cm}}}
\hline
\multicolumn{1}{c}{} & \multicolumn{15}{c}{\textbf{Mitigation Strategy}} \\
\hline
\textbf{Reference} &  
\rotatebox{54}{Differential Privacy} & 
\rotatebox{54}{Federated Learning} & 
\rotatebox{54}{SMPC} & 
\rotatebox{54}{xAI} & 
\rotatebox{54}{Homomorphic Encryption} & 
\rotatebox{54}{K-Anonymity} & 
\rotatebox{54}{L-Diversity} & 
\rotatebox{54}{Name-Entity Masking} & 
\rotatebox{54}{Data Aggregation} & 
\rotatebox{54}{Data Regularization} & 
\rotatebox{54}{Phonetic Normalization} & 
\rotatebox{54}{Topic Filtering} &
\rotatebox{54}{SHAP} & 
\rotatebox{54}{LIME} & \\

\hline
Identity Protection
~\cite{lohar2021irish,teodorescu2023language,leonardelli2021agreeing,hidayatullah2022systematic,berzak-etal-2017-predicting,malmasi2015arabic} &   &  &  & $\bullet$ &  &  &  &$\bullet$  & $\bullet$ &  &  & $\bullet$ &  &  \\
\hline
Model Explainability
~\cite{xiao2018mobile,mohamed2022artelingo,zampieri-etal-2024-federated,hidayatullah2022systematic,jauhiainen2019automatic,huang2015improved} & $\bullet$  & $\bullet$ &  &  &  &  &  & $\bullet$ & $\bullet$ & $\bullet$ &  & $\bullet$ &  & \\
\hline
Secured Computation
~\cite{arango2024multifold,xiao-etal-2024-toxicloakcn,mahmud2023cyberbullying,mehta2022social,gangarde2021privacy,alloghani2019systematic} &   &  & $\bullet$ & $\bullet$ & $\bullet$ & $\bullet$ & $\bullet$ & $\bullet$ & $\bullet$ & $\bullet$ & $\bullet$ & $\bullet$ & $\bullet$ & $\bullet$  \\
\hline
Data Refinement
~\cite{jeong-etal-2022-kold,sigurbergsson-derczynski-2020-offensive,berzak-etal-2017-predicting,ferragne2007automatic} & $\bullet$  & $\bullet$ &  &$\bullet$  &  &  &  &$\bullet$  & $\bullet$ & $\bullet$ & $\bullet$ & $\bullet$ &  &  \\
 \hline
 
\end{tabular}
}
\caption{Mitigation Strategies of Social Media Privacy Concerns}
\label{tab:result7}
\end{table*}

\end{document}